%% file: main.tex
\documentclass[10pt,twocolumn,letterpaper]{article}

\usepackage{iccv}
\usepackage{times}
\usepackage{epsfig}
\usepackage{graphicx}
\usepackage{amsmath}
\usepackage{amssymb}
\usepackage{booktabs}
\usepackage{url}
\usepackage{epstopdf}
\usepackage{cite}
\usepackage{gensymb}
\usepackage{url}
\usepackage[bottom]{footmisc}
\usepackage{enumitem}
\usepackage{caption}
\usepackage{float}
\usepackage{subcaption}
\usepackage{multirow} 
\usepackage{color}
\usepackage{accents}

\usepackage{array}
\newcolumntype{P}[1]{>{\centering\arraybackslash}p{#1}}
\usepackage{breqn}
\usepackage{tabularx}
\graphicspath{{./fig/}}

\usepackage{bm}

\newcommand*{\affaddr}[1]{#1} 
\newcommand*{\affmark}[1][*]{\textsuperscript{#1}}
\newcommand*{\email}[1]{\texttt{#1}}

\usepackage[pagebackref=true,breaklinks=true,letterpaper=true,colorlinks,bookmarks=false]{hyperref}

\iccvfinalcopy 

\def\iccvPaperID{6404} 

\ificcvfinal\fi
\begin{document}

\title{A Dual-Path Model With Adaptive Attention For Vehicle Re-Identification}

\author{%
Pirazh Khorramshahi\affmark[1], Amit Kumar\affmark[1], Neehar Peri\affmark[1], Sai Saketh Rambhatla\affmark[1], Jun-Cheng Chen\affmark[2] \\ and Rama Chellappa\affmark[1]\\
\affaddr{\affmark[1]Center for Automation Research, UMIACS, University of Maryland, College Park}\\
\affaddr{\affmark[2]Research Center for Information Technology Innovation, Academia Sinica}\\
\email{\tt\small \{pirazhkh, akumar14, peri, rssaketh, rama\}@umiacs.umd.edu}, \ \email{\tt\small pullpull@citi.sinica.edu.tw} 
}



\maketitle
%

\input{abstract.tex}
\input{intro.tex}

\input{related.tex}

\input{method.tex}
\input{experiment.tex}

\input{exp.tex}

\input{conclusion.tex}
\input{acknowledgement.tex}


{

\input{main.bbl}
}

\input{supp.tex}

\end{document}

%% file: abstract.tex
\begin{abstract}
In recent years, attention models have been extensively used for person and vehicle re-identification. Most re-identification methods are designed to focus attention on key-point locations. However, depending on the orientation, the contribution of each key-point varies. In this paper, we present a novel dual-path adaptive attention model for vehicle re-identification (AAVER). The global appearance path captures macroscopic vehicle features while the orientation conditioned part appearance path learns to capture localized discriminative features by focusing attention on the most informative key-points. Through extensive experimentation, we show that the proposed AAVER method is able to accurately re-identify vehicles in unconstrained scenarios, yielding state of the art results on the challenging dataset VeRi-776. As a byproduct, the proposed system is also able to accurately predict vehicle key-points and shows an improvement of more than $7\%$ over state of the art. The code for key-point estimation model is available at \small{\url{https://github.com/Pirazh/Vehicle_Key_Point_Orientation_Estimation}}
\end{abstract}

%% file: intro.tex
\section{Introduction}\label{sec:intro}
Vehicle re-identification (re-id) refers to the task of retrieving all images of a particular vehicle identity in a large gallery set, composed of vehicle images taken from varying orientations, cameras, time and locations. Accurately re-identifying vehicles from images and videos, is of great interest in surveillance and intelligence applications. In contrast to vehicle recognition which aims to identify the make and model of the vehicle, vehicle re-id is concerned with identifying specific vehicle instances. This task is extremely challenging as vehicles with different identities can be of the same make, model and color, and thus it is challenging for a Deep Convolutional Neural Network (DCNN) to make accurate predictions. In this paper, we present a novel algorithm driven by adaptive attention for re-identifying vehicles from still images without using information from other sources such as time and location.

The similar task of person re-id aims at re-identifying humans appearing in different cameras. While visual appearance models work reasonably well for person re-id, the same techniques fail to differentiate vehicles due to the lack of highly discriminating features. Person re-id models are not heavily reliant on facial features as they also learn discriminating features based on clothing and accessories. However, vehicle re-id poses a new set of challenges. Different vehicle identities can have similar colors and shapes especially those coming from the same manufacturer with a particular model, trim and year. Subtle cues such as different wheel patterns and custom logos might be unavailable in the global appearance features. Therefore, it is important that vehicle re-id model learns to focus on different parts of the vehicles while making a decision. Previous works in person re-id such as \cite{Xu_2018_CVPR} have used attention models with human key-points as regions of attention and have shown significant improvement in performance. Similarly, methods such as \cite{wang2017orientation} have used vehicle key-points to learn attention maps for each of the $20$ key-points defined by \cite{wang2017orientation}. The system proposed by Wang \emph{et al.}\cite{wang2017orientation} grouped key-points into four groups corresponding to front, rear, left and right. 

\begin{figure}[t]
\centering
    \begin{subfigure}[t]{0.075\textwidth}
    \centering

    \includegraphics[height=1.35cm,width=1.35cm]{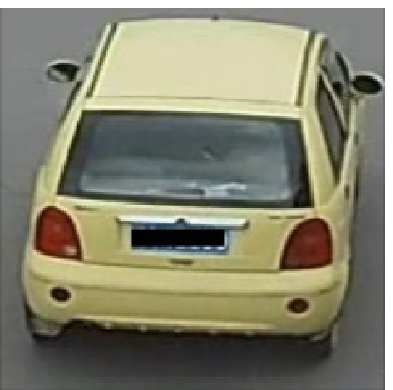}
    
    \end{subfigure}
    \begin{subfigure}[t]{0.075\textwidth}
    \centering
    \includegraphics[height=1.35cm,width=1.35cm]{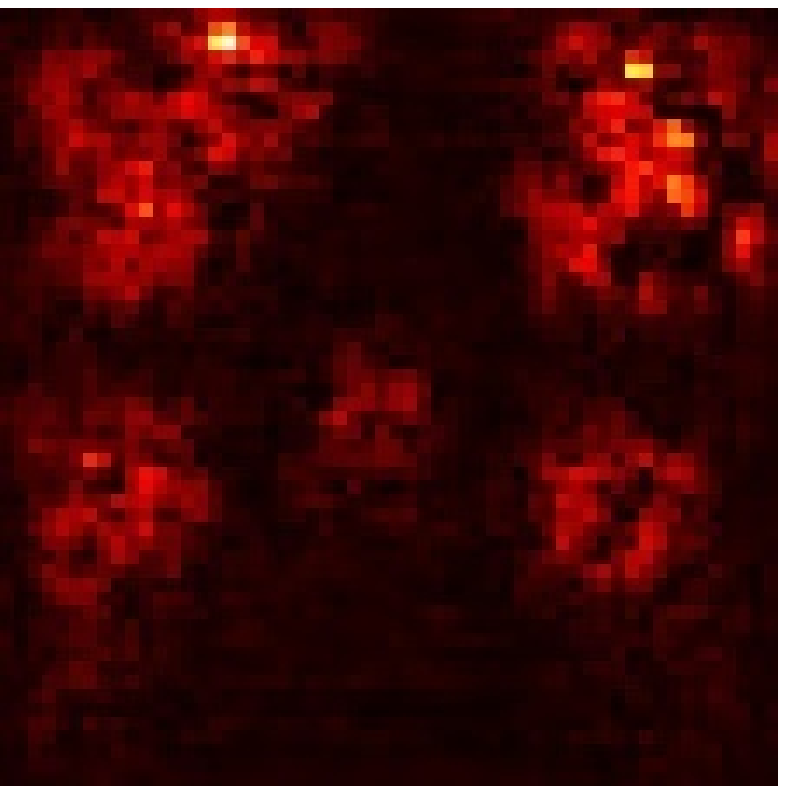}
    \caption{Front}
    \label{fig:fig_1a}

    \end{subfigure}
    \begin{subfigure}[t]{0.075\textwidth}
    \centering
    \includegraphics[height=1.35cm,width=1.35cm]{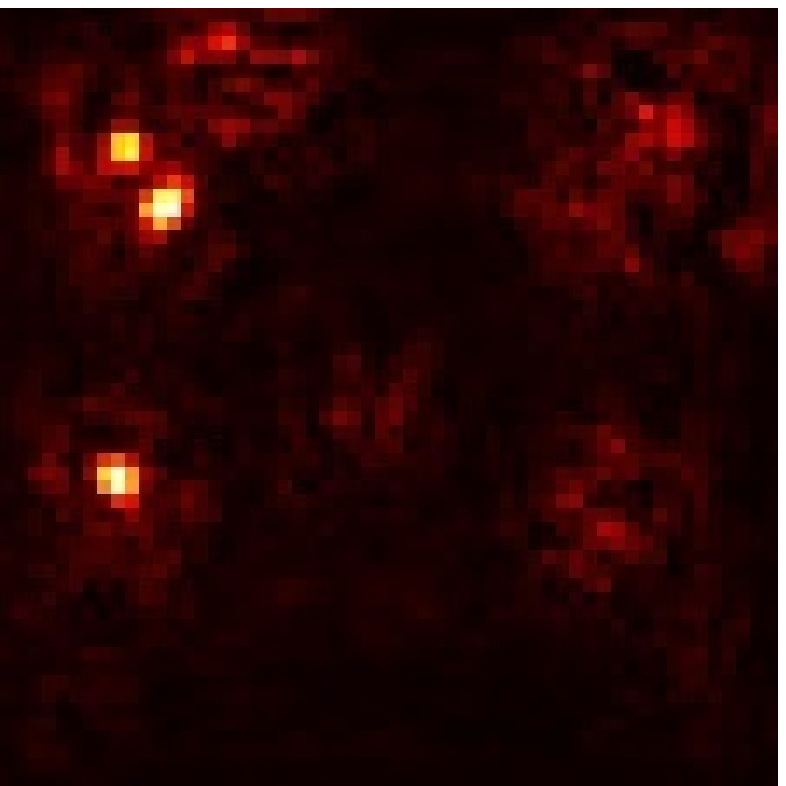}
    \caption{Left}
    \end{subfigure}
    \begin{subfigure}[t]{0.075\textwidth}
    \centering
    \includegraphics[height=1.35cm,width=1.35cm]{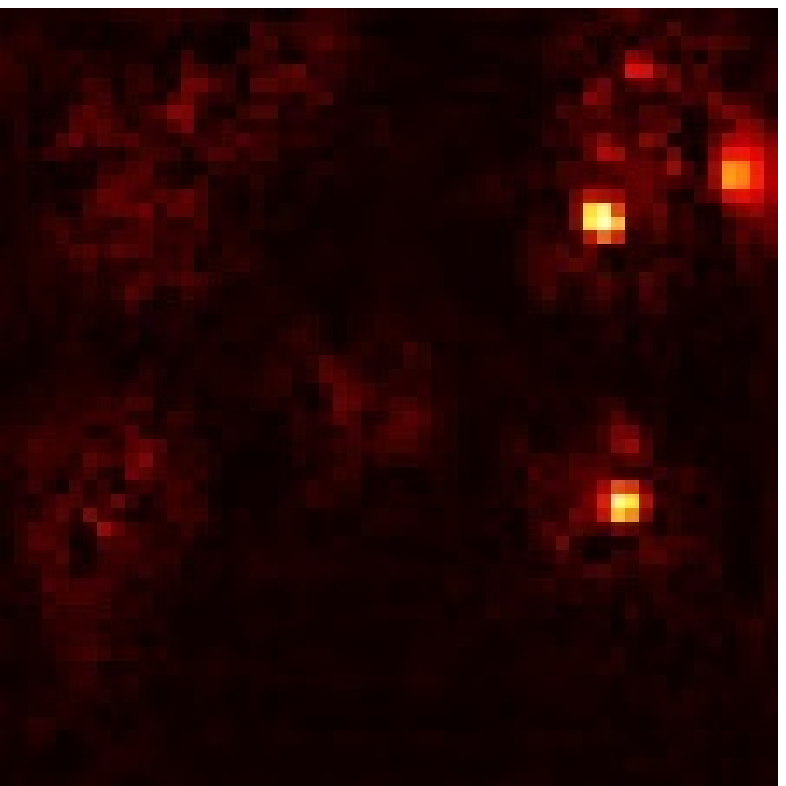}
    \caption{Right}
    \end{subfigure}
    \begin{subfigure}[t]{0.075\textwidth}
    \centering
    \includegraphics[height=1.35cm,width=1.35cm]{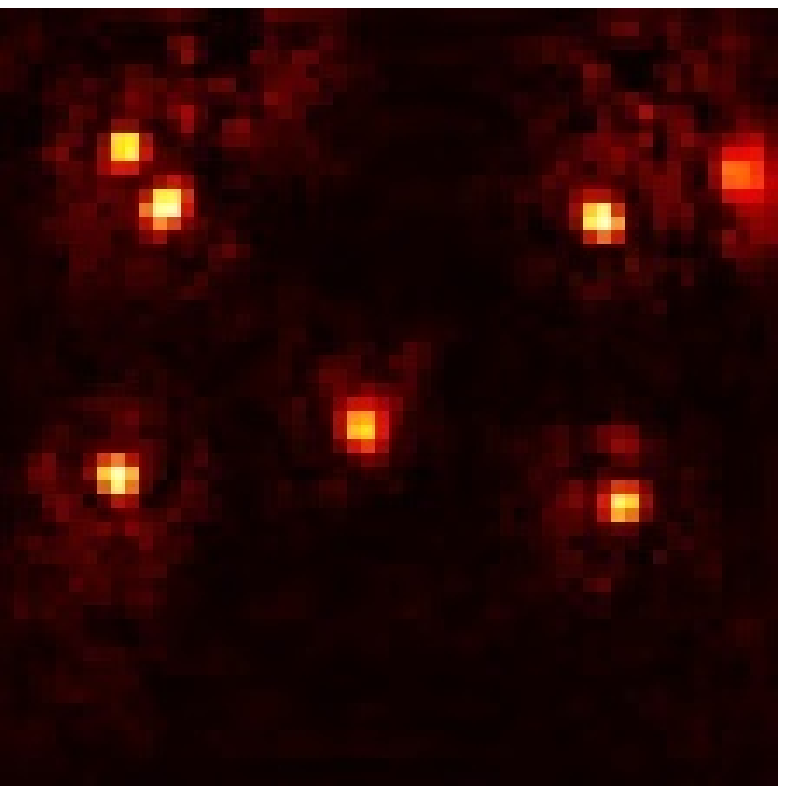}
    \caption{Rear}
    \end{subfigure}
       
    \caption{Heatmaps grouped as suggested in \cite{wang2017orientation}. Attention to 
    all subgroup of key-points leads to erroneous results. Although, only the rear of the car is visible, contributions from frontal key-points are non-zero.}
     \label{fig:fig_1} 
\end{figure}


However, not all key-points provide discriminating information and their respective contributions depend on the orientation of the vehicle. For instance, in Figure \ref{fig:fig_1a} we observe that the key-points from the front of the car incorrectly influence the attention of the model as the front of the car is not visible. Hence, paying attention to all the key-points, as suggested in \cite{wang2017orientation}, can lead to erroneous results. The proposed method tackles the problem of false attention by adaptively choosing the key-points to focus on, based on the orientation of the vehicle hence, providing complementary information to global appearance features. In this work, terms with same connotation, \textit{path}, \textit{stream} and \textit{branch}, have been used interchangeably.  

In the proposed method, the first stream is a DCNN trained to extract discriminative global appearance features for each vehicle identity. However, this stream often fails to extract subtle features necessary to distinguish similar vehicles. Therefore, a second path composed of orientation conditioned key-point selection and localized feature extraction modules is used in parallel to supplement the features from the first path. By using orientation as a conditioning factor for adaptive key-point selection, the model learns to focus on the most informative parts of the vehicle. Additionally, we develop a fully convolutional two-stage key-point detection model inspired by the works of Kumar \emph{et al.}\cite{Kumar_2018_CVPR} and Bulat \emph{et al.}\cite{bulat2016human} for facial key-point detection and human pose estimation respectively.  

The detailed architectures of each module in the proposed method are discussed in section \ref{sec:proposed}. Through extensive experimentation, we show that the proposed Adaptive Attention model for Vehicle Re-identification (AAVER) approach improves the re-id accuracy on challenging datasets such as VeRi-776 \cite{liu2016large,liu2016deep_Veri} and VehicleID \cite{liu2016deep_VehicleID}. In addition, the proposed vehicle key-point detection model, improves the accuracy by more than $7\%$ over state of the art.

%% file: related.tex
\section{Related Work}\label{sec:related}

\input{jc.tex}

%% file: jc.tex

 
In this section, we briefly review recent relevant works in the field of vehicle classification and re-identification.
Learning a discriminating representation, requires a large-scale annotated data for training, especially for recent DCNN approaches. Yang \emph{et al.}\cite{yang2015large} released a large-scale car dataset (CompCars) for fine-grained vehicle model classification which consists of 1,687 car models and 214,345 images. The VehicleID dataset by Liu \emph{et al.} \cite{liu2016deep_VehicleID} consists of 200,000 images of about 26,000 vehicles. In addition, Liu \emph{et al.}\cite{liu2016deep_Veri,liu2018provid} published a high-quality multi-view vehicle re-id (VeRi-776) dataset. Yan \emph{et al.}\cite{yan2017exploiting} released two high-quality and well-annotated vehicle datasets, namely VD1 and VD2, with diverse annotated attributes, containing 1,097,649 and 807,260 vehicle images captured in different cities.

Moreover, besides datasets for training, Tang \emph{et al.}\cite{tang2017multi} claimed
 traditional hand-crafted features are complementary to deep features and thus fused both features to realize an improved
 representation. Instead, Cui \emph{et al.}\cite{cui2017vehicle} fused features from  various DCNNs trained with different objectives. Furthermore, Liu \emph{et al.}\cite{liu2016deep_Veri,liu2018provid} used multi-modal features, including visual features, license plate, camera location, and other contextual information, in a coarse-to-fine vehicle retrieval framework. To augment the training data for robust training,  \cite{wu2018rerank} used a generative adversarial network to synthesize vehicle images with diverse orientation and appearance variations. \cite{zhou2018viewpoint} learns a viewpoint-aware representation for vehicle re-id through adverserial learning and a viewpoint-aware attention model. 
 
 Besides global features, Liu \emph{et al.}~\cite{liu2018ram} extracted discriminative local features from a series of local regions of a vehicle by a region-aware deep model. Different from these approaches, the proposed method leverages orientation to adaptively select the regions of attentions.


Another effective strategy to learn the discriminative representation is metric learning. Zhang \emph{et al.}\cite{zhang2017improving} proposed an improved triplet loss which performs joint optimization with an auxiliary classification loss as a regularizer in order to characterize intra-sample variances. Bai \emph{et al.}\cite{8265213} introduced Group-Sensitive triplet embeding to better model the intra-class variance. Shen \emph{et al.}~\cite{shen2017learning} also proposed to improve the matching performance by making use of spatio-temporal information; they developed a Siamese-CNN with path LSTM model which generates the corresponding candidate visual-spatio-temporal paths of an actual vehicle image by a chain-based Markov random field (MRF) model with a deeply learned potential function.
In contrast, the proposed method uses the $L_2$ softmax \cite{ranjan2017l2} loss function as it has shown impressive performance for the task of face verification and trains faster compared to triplet loss-based methods such as \cite{zhang2017improving} without the hassle of sampling hard triplets.  



%% file: method.tex
\section{Adaptive Attention Vehicle Re-identification (AAVER)}
\label{sec:proposed}

\begin{figure*}[ht]
\centering
 \includegraphics[width=\linewidth]{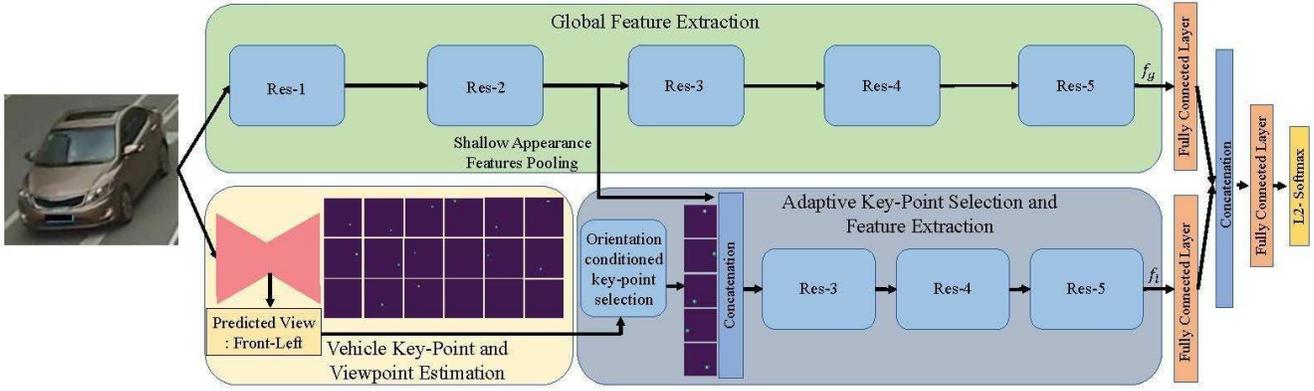}
  \caption{Adaptive Attention Vehicle Re-identification (AAVER) Model Pipeline. The input vehicle image is processed in parallel along two paths: In the first path, the global appearance features ($f_g$) are extracted. The second path is responsible for detecting vehicle key-points and predicting its orientation, after which localized features ($f_l$) are extracted based on adaptive key-point selection. Subsequently, the two feature vectors $f_g$ and $f_l$ are fused with a shallow multi-layer perceptron.}
  \label{fig:partmodel}
\end{figure*}

The entire pipeline of the proposed method AAVER is composed of three main modules: \textbf{Global Feature Extraction}, \textbf{Vehicle Key-Point and Viewpoint Estimation}, and \textbf{Adaptive Key-Point Selection and Feature Extraction} which is followed by a re-ranking based post-processing. Figure \ref{fig:partmodel} shows the diagrammatic overview of our method.

In AAVER, the global feature extraction module is responsible for extracting the macroscopic features ($f_g$) of the vehicles. By looking at the entire vehicle, this model tries to maximally separate the identities in the feature space. However, this model may fail to take into account subtle differences between similar cars, most extremely the ones that are of the same make, model and color. Therefore, the features generated by this module are supplemented with features ($f_l$) from the localized feature extraction module. This can be achieved by the adaptive attention strategy using the proposed key-point and orientation estimation network.

In order to estimate the vehicle key-points, we draw inspiration from literature on facial key-point detection and human pose estimation. Inspired by \cite{Kumar_2018_CVPR,bulat2016human} we employ a two-stage model to predict the vehicle's orientation and landmarks in a coarse to fine manner; the coarse heatmaps predicted by a DCNN are refined using a shallower network.

Finally, we use the proposed adaptive key-point selection module to select a subset of most informative key-points and pool features from early layers of the global feature extraction module to extract localized features around the selected key-points. The features obtained from the two paths of AAVER are then merged using a multi-layer perceptron. The entire model can be trained end-to-end using any differentiable loss function. In our work, we use the $L_2$ softmax loss as proposed in \cite{ranjan2017l2}.
During inference, we use the features from  penultimate fully connected layer as the representation of a given vehicle. Additionally, we also perform re-ranking\cite{zhong2017re} as a post processing step.

Each module is described in detail in the following sub sections. Pytorch deep learning framework \cite{paszke2017automatic} has been used in all of the experiments.

\input{pirazh.tex}

\input{saket.tex}

%% file: pirazh.tex
\begin{figure*}
    \centering
    \includegraphics[width=0.82\textwidth]{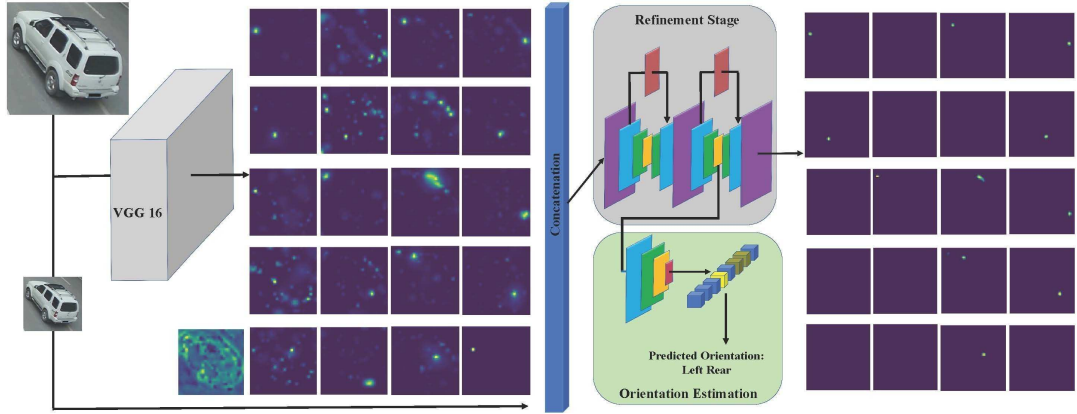}
    \caption{Vehicle key-point and orientation estimator network. VGG $16$ network outputs $21$ coarse heatmaps corresponding to the 20 vehicle landmarks and the background (Response maps on the left). A two-stack hourglass network refines $20$ key-points heatmaps (response maps on the right) excluding background channel and predicts the vehicle's orientation.}
    \label{fig:KP_net}
\end{figure*}

\begin{figure}
    \centering
 \includegraphics[width=0.45\textwidth]{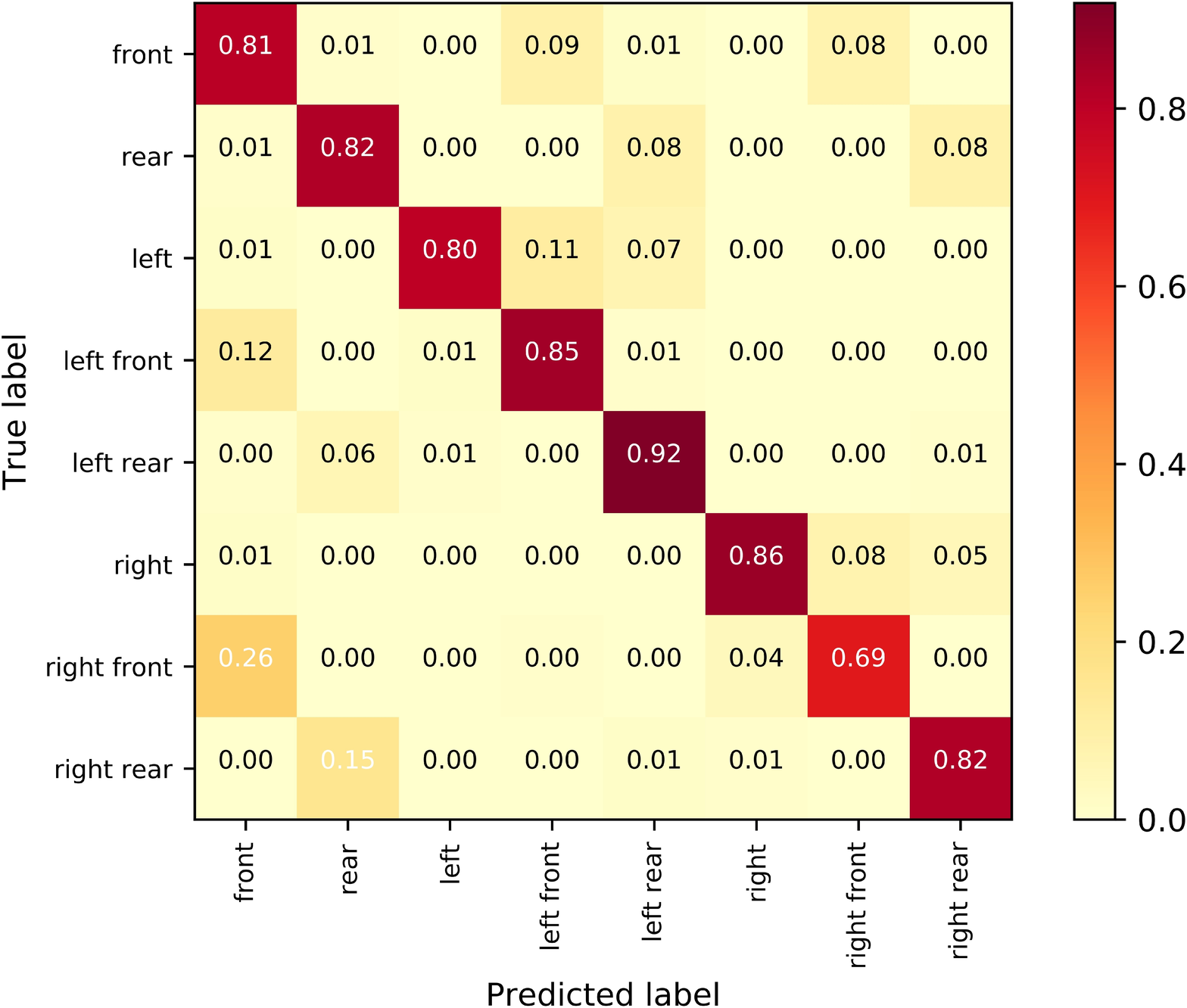}
  \caption{The confusion matrix of the vehicle orientation estimation network}
  \label{fig:pose_confusionmatrix}
\end{figure}

\subsection{Global Feature Extraction} \label{subsec:global}

For extracting the global appearance features, we employ ResNet-50 and ResNet-101 \cite{DBLP:journals/corr/HeZRS15} as backbone networks and also adopt them as our baseline models. We initialized the weights of these models using the weights from the models pre-trained on the CompCars dataset. A $2048$-dimensional features vector from the last convolutional layer of ResNet is then fed to a shallow multi-layer perceptron. This network is trained using the $L_2$ softmax loss function which constrains the feature vectors extracted by the network to lie on a hyper-sphere of radius $\alpha$. This enables the network to embed features of identical vehicles together while pushing apart the features from different vehicles. It is mathematically expressed as:
\begin{equation}
    \label{eqn:l2softmax}
    \centering
    \mathcal{L}_S = - \log \frac{\exp^(\mathbf{W}_{y}^T (\frac{\alpha \mathbf{x}}{\| \mathbf{x} \|_2}) + b_{y})}{\sum_{j = 1}^{N} \exp^(\mathbf{W}_j^T (\frac{\alpha \mathbf{x}}{\| \mathbf{x} \|_2}) + b_j)}
\end{equation}
\noindent
where $\mathbf{x}$ is the feature vector corresponding to class label $y$, $\mathbf{W}_j$ is the weight and $b_j$ is the bias corresponding to class j, $\alpha$ is a positive trainable scalar parameter, and $N$ is the number of classes respectively.

\subsection{Vehicle Key-Point and Orientation Estimation}

In this work, a two-stage model is proposed for key-point estimation. In the first stage, a VGG-16\cite{simonyan2014very} -based fully convolutional network is employed to do a coarse estimation of the location of $N_1$ ($N_1=21=$ 20 key-points plus background) heatmaps of size $H \times W$ ($56 \times 56$). This network is trained using a per-pixel multi-class cross entropy loss defined as follows:
\begin{equation}
    \centering
    \mathcal{L}_1 =  \frac{-1}{H \times W} \sum_{i=1}^{H}\sum_{j=1}^{W}\log(\frac{\exp(\mathbf{x}_{i,j}({t^{*}}_{i,j}))}{\sum_{k=1}^{N_1}\exp(\mathbf{x}_{i,j}(k))})
    \label{eqn:kpnet_stage1_loss}
\end{equation}
\noindent
where $\mathbf{x}_{i,j}$ is the vector corresponding to pixel location $i$ and $j$ across all output channels and $t^{*}_{i,j}$ is the ground-truth class label for that pixel location.  After training the first stage, the weights of this network are frozen for training of the subsequent stage. The left side of Figure \ref{fig:KP_net} depicts the the output of the first stage for a sample vehicle image. 

Although the responses of the first stage can be used for the prediction of visible key-point locations, there might be erroneous activations in the heatmaps that correspond to invisible key-points. Consequently, we use the second stage that takes in the sub-sampled version of the input image and the coarse estimates of key-points to refine the results. The refinement network follows the hourglass architecture introduced in \cite{newell2016stacked} which is commonly used for refining heatmaps and reducing artifacts due to invisible key-points. In the second stage, coarse heatmaps estimated from the first stage, are refined through a two-stack hourglass network with skip connections. Along with refining the estimated key-points, the orientation of the vehicle is also predicted through a parallel branch composed of two fully connected layers designed to classify the orientation into eight classes as defined in\cite{wang2017orientation}. This multi-task learning helps the refinement network to make accurate predictions of the visible key-points while reducing the response of invisible key-points. Figure \ref{fig:KP_net} shows the overall schematic flow of the two-stage network.

To train the heatmap refinement and orientation branches we use Mean Square Error (MSE) and cross entropy loss respectively. Equation \ref{eqn:kpnet_stage2_loss} represents the loss function used for the second stage. It is worth mentioning that in the second stage we are only interested in foreground heatmaps, hence, we exclude the refinement of the background channel.

\begin{equation}
    \centering
    \mathcal{L}_2 = \mathcal{L}_H + \lambda * \mathcal{L}_O
    \label{eqn:kpnet_stage2_loss}
\end{equation}
where $\mathcal{L}_H$ is the heatmap regression loss:
\begin{equation}
    \mathcal{L}_H = \sum_{k=1}^{N_2}\sum_{i=1}^{H}\sum_{j=1}^{W} {\mid h_{k}(i,j) - {h^{*}}_{k}(i,j)\mid}^2
    \label{eqn:kpnet_stage2_Heatmaploss}
\end{equation}
and $\mathcal{L}_O$ is the orientation classification loss:
\begin{equation}
    \mathcal{L}_O = -\log(\frac{\exp(\mathbf{p}(p^{*}))}{\sum_{i=1}^{N_p} \exp(\mathbf{p}(i))}).
    \label{eqn:kpnet_stage2_poseloss}
\end{equation}

In Equation (\ref{eqn:kpnet_stage2_Heatmaploss}), $N_2 = N_1 -1$, $h_{k}(i,j)$ and $h^{*}_{k}(i,j)$ are predicted and ground-truth heatmaps in stage 2 for the $k^{th}$ key-point at locations $i$ and $j$ respectively. $\mathbf{p}$, $p^{*}$ and $N_p$ in Equation (\ref{eqn:kpnet_stage2_poseloss}) constitute the predicted orientation vector, the corresponding ground-truth orientation and number of classes respectively. Finally, $\lambda$ in Equation (\ref{eqn:kpnet_stage2_loss}) is a weight to balance the losses used in model optimization. In our experiments, $\lambda$ is set to $10$ obtained after cross-validation. In the right hand side of Figure \ref{fig:KP_net} which shows the the output of the second stage, it can be observed that the initial coarse estimates of key-points have been refined.

\subsection{Adaptive Key-Point Selection and Feature Extraction}

 Subtle differences in similar vehicles mostly occur close to vehicle landmarks, \emph{e.g.} same car make and models of same color might be distinguishable through their window stickers, rims, indicator lights on the side mirrors, etc. This can be achieved by focusing the attention on parts of the image that encompasses these distinctions. To this end, regions of interest within the image are identified based on the orientation of the vehicle; after which features from the shallower layer of the global appearance model are pooled. As suggested in \cite{DBLP:journals/corr/ZeilerF13} these pooled features contain contextual rather than abstract information. Later, deep blocks (Res3, Res4 and Res5) of another ResNet model are used to extract supplementary features corresponding to the regions of interest. 

In \cite{wang2017orientation}, vehicle's orientation is annotated into eight different classes, \emph{i.e.} rear, left ,left front, left rear, right, right front and right rear; however, there is no absolute boundary between two adjacent orientations. For instance, for the case of right and right front, the network gets confused between the two classes when trained for orientation prediction; this can be observed in Figure \ref{fig:pose_confusionmatrix} which shows the confusion matrix for the eight-class classification problem. To overcome this issue, we designed a key-point selector module that takes the predicted orientation likelihood vector and adaptively selects the key-points based on the likelihoods. 

In order to achieve this, we constructed eight groups shown in Table \ref{tab:KP_selection_map} corresponding to each of the eight orientations of a vehicle and its two adjacent orientations. During inference, the likelihood of each orientation group is calculated and the one with the highest probability is picked. Also, experimentally it was observed that for each orientation group at least seven key-points are always visible. Consequently, given the orientation group with the highest probability we select the seven heatmaps shown in Table \ref{tab:KP_selection_map} corresponding to the respective orientation group. These orientation groups are named based on their center orientation \emph{e.g.} the group that contains left front, front and right front is named front. 

\begin{table}
    \centering
    \caption{Seven Prominent key-points in each orientation group}
    \resizebox{\columnwidth}{!}{%
    \begin{tabular}{|c|P{5cm}|} 
    \hline
    Orientation Group & Visible Key-Points  \\ 
    \hline\hline
    Front        &     [11, 12, 7, 8, 9, 13, 14]     \\ 
    \hline
    Rear        & [18, 16, 15, 19, 17, 11, 12]        \\ 
    \hline
    Left         & [8, 1, 11, 14, 15, 2, 17]        \\ 
    \hline
    Left Front    & [9, 14, 6, 8, 11, 1, 15]        \\ 
    \hline
    Left Rear        & [2, 17, 15, 11, 14, 19, 1]        \\ 
    \hline
    Right        & [7, 3, 12, 13, 16, 4, 18]        \\
    \hline
    Right Front &  [9, 13, 5, 7, 12, 3, 16] \\
    \hline
    Right Rear &   [3, 4, 12, 16, 18, 19, 13] \\
    \hline
    \end{tabular}
    }
    \label{tab:KP_selection_map}
\end{table}

After obtaining the seven heatmaps, for each map, a Gaussian kernel with $\sigma=2$ is placed in the location of the map's peak, \emph{i.e.} the key-point location. This is done in order to emphasize the importance of the surrounding areas around the key-points as they may have discriminative information. 

Following the adaptive heatmap selection and dilation by the Gaussian kernel, is the localized feature extraction ($f_l$) by Res3, Res4 and Res5 blocks of the parallel ResNet model. The input to this sub-network is the concatenation of the seven dilated heatmaps of shape $7 \times 56 \times 56$ and the pooled global features of shape $256 \times 56 \times 56$. Finally, the localized features $f_l$ is concatenated with the global appearance features $f_g$ and passed through a multi-layer perceptron followed by $L_2$ softmax loss function (refer to Figure \ref{fig:partmodel}). Given that features are normalized, we use cosine similarity to calculate the similarity score between image pairs.

\subsection{Post Processing Step: Re-Ranking}\label{reranking}
In general, Re-ID can be regarded as a retrieval problem. Given a probe vehicle, we want to search in the gallery for images containing the same vehicle in a cross-camera mode. After an initial ranking list is obtained, a good practice consists of adding a re-ranking step, with the expectation that the relevant images will receive higher ranks. Such re-ranking steps have been mostly studied in generic instance retrievals such as \cite{qin2011hello}, \cite{chum2007total}, \cite{jegou2007contextual} and \cite{zhong2017re}.
 The main advantage of many re-ranking methods is that they can be implemented without requiring additional training samples, and also can be applied to any initial ranking list.

Significant amount of research in person re-id goes into re-ranking strategies and vehicle re-id is lacking in that aspect. Most of the state of the art methods for vehicle re-id do not perform re-ranking on their initial ranking list. 
We use the re-ranking strategy proposed by Zhong et al. \cite{zhong2017re} in our work.

%% file: experiment.tex
\section{Experiments}
\label{sec:experiments}
Here we first present the two large-scale datasets used for the vehicle re-identification task and their evaluation protocols, after which we describe the implementation details of the proposed method. 

\subsection{Datasets}\label{subsec:datasets}
To the best our knowledge, there are mainly two large scale vehicle datasets that are publicly available and are designed for the task of vehicle re-identification: VeRi-776 \cite{liu2016large},\cite{liu2016deep_Veri} and VehicleID \cite{liu2016deep_VehicleID}.

\textbf{VeRi-776} dataset consists of 49,357 images of $776$ distinct vehicles that were captured with 20 non-overlapping cameras in variety of orientations and lighting conditions. Out of these images, 37,778 (576 identities) and 11,579 (200 identities) have been assigned to training and testing respectively. For the query set, 1,678 images have been selected from the testing set. The evaluation protocol for this dataset is as follows: for each probe image in the query set the corresponding  identity and the camera ID from which the image is captured is gathered. The gallery is constructed by selecting all the images in the testing set except the ones that share the same identity and camera ID as the probe. Evaluation metrics adopted for this dataset are mean Average Precision (mAP), Cumulative Match Curve (CMC) for top 1 (CMC@1) and top 5 (CMC@5) matches.

\textbf{VehicleID} is another large-scale dataset used for vehicle retrieval task and is composed of $221,567$ images from $26,328$ unique vehicles. Half of the identities, \emph{i.e.} $13,164$,  are reserved for training while the other half are dedicated for evaluation. There are $6$ test splits for gallery sizes of $800$, $1600$, $2400$, $3200$, $6000$ and $13,164$. In the recent works \cite{wang2017orientation,shen2017learning} the first three splits have been used. The proposed evaluation protocol for each split in VehicleID dataset is to randomly select an image for each of the identities to form the gallery of respective size and use the rest of the images for query. This procedure is repeated ten times and the averaged metrics, CMC@1 and CMC@5, are reported. 

\subsection{Implementation Details}

In our implementation, all the input images were resized to $(224,224)$ and normalized by the ImageNet dataset \cite{deng2009imagenet} mean and standard deviation. Also, in all of our experiments we used batch training with size of $150$ and Adam optimizer \cite{kingma2014adam} with the learning rate of $1e-4$.

Initially, we fine-tuned our baseline models (see section \ref{subsec:global}) on VeRi-776 and VehicleID datasets separately, for $20$ epochs. Then, we initialized the key-point and orientation estimation network with ImageNet pre-trained weights. The first stage of this network was trained for $40$ epochs; afterwards the second stage was trained for $40$ epochs as well. 

Next, we trained the orientation conditioned feature extraction branch for each of VeRi-776 and VehicleID datasets for 20 epochs. Finally, we select the network's output of the penultimate layer as the feature vector corresponding to the input vehicle image.

%% file: exp.tex
\section{Experimental Evaluations}\label{sec:evaluation_results}

We first present the evaluation results of our vehicle key-point and orientation estimation model followed by the evaluation of the proposed method AAVER on both VeRi-776 and VehicleID datasets.

\subsection{Vehicle Key-Point and Orientation Estimation Evaluation}

In order to evaluate the performance of the proposed two-stage key-point detection model, we use the Mean Square Error (MSE) in terms of pixels for the location of visible key-points in $56 \times 56$ maps over the test set of VeRi-776 key-point dataset. Table \ref{tab:kp_net_acc} shows the MSE of our model after first and second stages. Moreover, we measured the accuracy of the model for viewpoint classification. It can be observed that the refinement stage reduces the key-point localization error by $20\%$ compared to the first stage.  

To the best of our knowledge, \cite{wang2017orientation} is the only work on the VeRi-776 key-point and orientation estimation dataset. \cite{wang2017orientation} used the averaged distance between estimated and ground-truth locations of all visible key-points for evaluation. If the distance is less than a threshold value ($r_0$ in terms of pixels in $48 \times 48$ map), the estimation is considered to be correct. We follow the same protocol to compare the precision with \cite{wang2017orientation} and Table \ref{tab:kp_net_acc} shows the result of this comparison.

\begin{table}[h!]
    \centering
    \caption{Accuracy evaluation and comparison of the vehicle landmark and orientation estimation network}
\begin{tabular}{|c|c|c|} 
\cline{2-3}
\multicolumn{1}{c|}{} & Stage 1 & Stage 2  \\ \hline
Key-point localization MSE (pixel) & 1.95 & \textbf{1.56} \\ 
\hline
Orientation Accuracy & - & 84.44\%   \\ 
\hline
\hline
\multicolumn{3}{|c|}{Key-Point Precision Comparison} \\
\hline
\hline
Model & $r_0 = 3$ & $r_0 = 5$ \\
\hline
OIFE \cite{wang2017orientation} & 88.8\% & 92.05\%\\
\hline
Ours & \textbf{95.30\%} & \textbf{97.11\%} \\
\hline
\end{tabular}
    
    \label{tab:kp_net_acc}
\end{table}

\subsection{Evaluation Results on VeRi-776}

 Table \ref{tab:baseline_proposed_comparison} summarizes the results of the global appearance model (baseline) and the proposed AAVER model with adaptive attention. Note that in both ResNet-50 and ResNet-101 -based architectures, there is a significant improvement in mAP and CMC@1 scores after incorporating adaptive attention. This indicates that conditioning on the orientation of the vehicle and selecting corresponding key-points enables the network to focus more on parts that contains minute differences in similar cars. This claim is further studied in section \ref{subsec:ablation}. Unsurprisingly, we also observe that ResNet-101 shows better performance compared to ResNet-50 under similar settings.

\begin{table}[!htb]
    \centering
    \caption{Performance comparison between baseline and the proposed method on VeRi-776 dataset}
    \resizebox{\columnwidth}{!}{%
    \begin{tabular}{|c|c|c|c|c|}
    \hline
    \multicolumn{2}{|c||}{Model} & mAP & CMC@1 & CMC@5  \\
    \hline
    \multicolumn{1}{|c}{\multirow{2}{*}{Baseline}} & \multicolumn{1}{|c||}{ResNet-50} & 52.88  & 83.49  & 92.31   \\
    \cline{2-5}
    \multicolumn{1}{|c}{} & \multicolumn{1}{|c||}{ResNet-101} & 55.75  & 84.74  & 94.34  \\
    \hline
    \hline
    \multicolumn{1}{|c}{\multirow{2}{*}{AAVER}} & \multicolumn{1}{|c||}{ResNet-50} & 58.52  & 88.68  & 94.10  \\
    \cline{2-5}
    \multicolumn{1}{|c}{} & \multicolumn{1}{|c||}{ResNet-101} & \textbf{61.18}  & \textbf{88.97} & \textbf{94.70} \\
    \hline
    \end{tabular}
    }
    \label{tab:baseline_proposed_comparison}
\end{table}

\begin{figure}
    \centering
    \begin{tabular}[t]{cc}
        \begin{subfigure}{0.16\textwidth}
            \includegraphics[width=\textwidth,height=\textwidth]{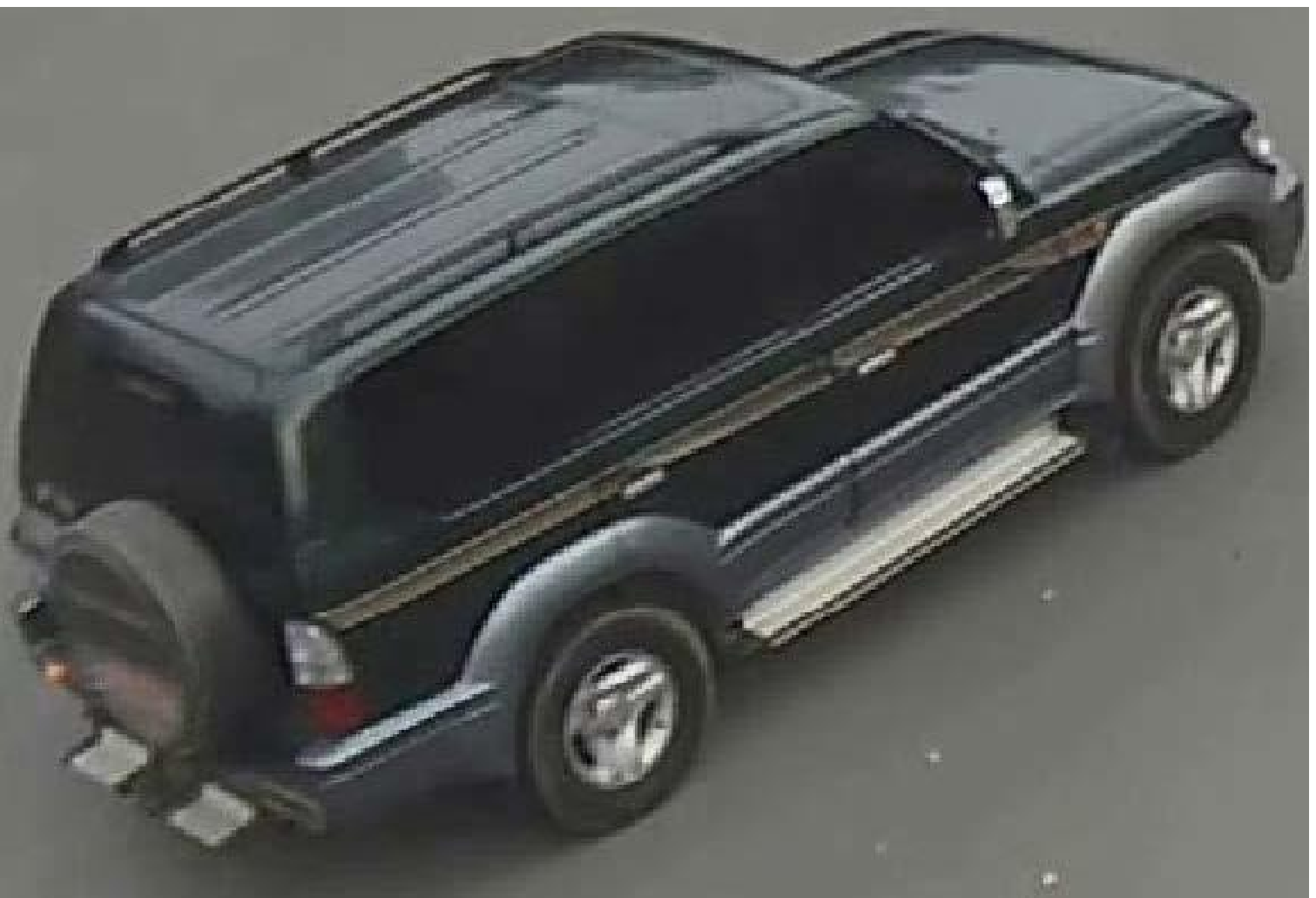}
            \caption{Probe Image}
        \end{subfigure}
        \hspace{-0.2cm}
        &  
        \begin{tabular}{ccc}
            \begin{subfigure}[t]{0.08\textwidth}
                \includegraphics[width=\textwidth]{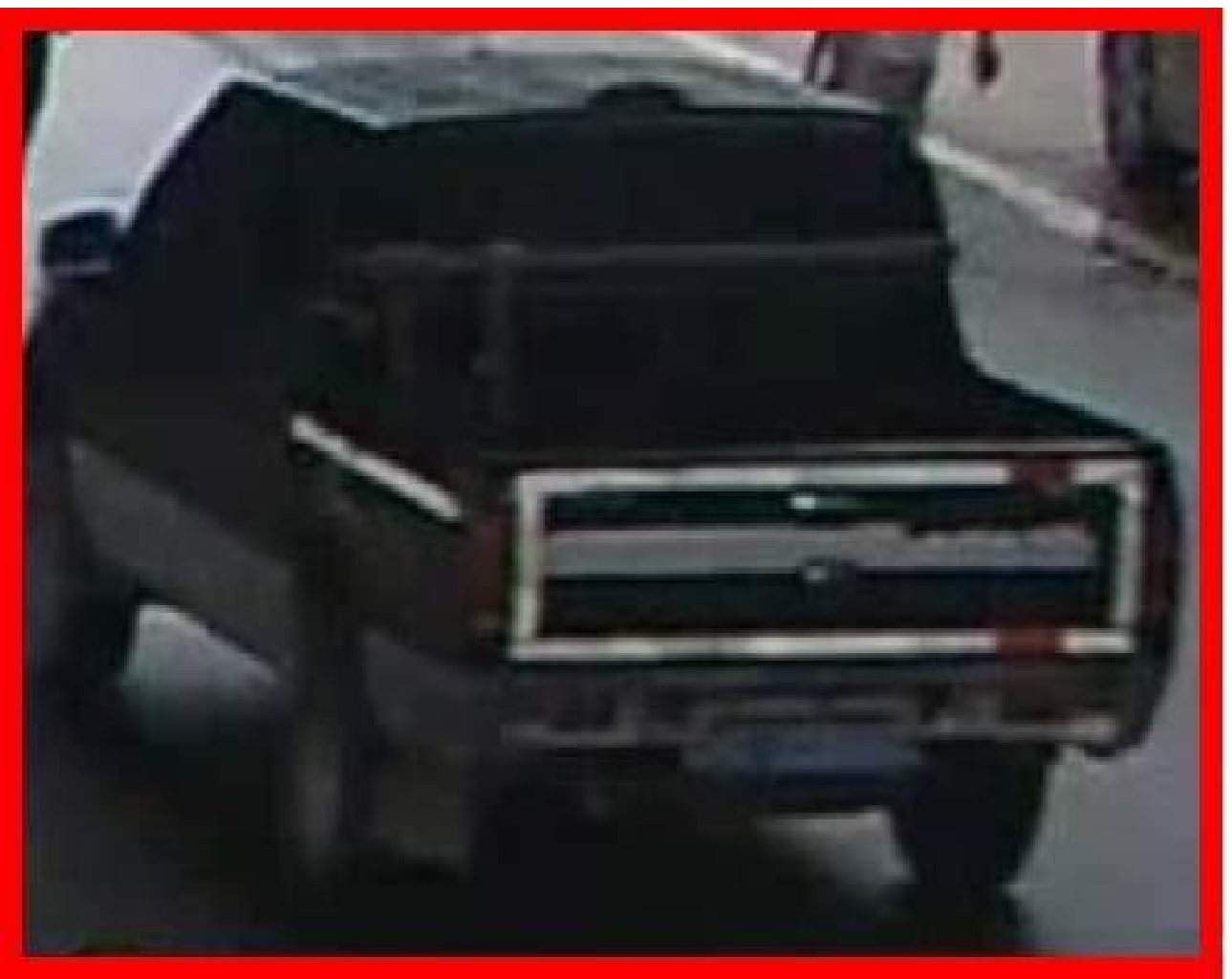}
                \caption{Rank $1$}
            \end{subfigure}
            \hspace{-0.2cm}
            &
            \begin{subfigure}[t]{0.08\textwidth}
                \includegraphics[width=\textwidth]{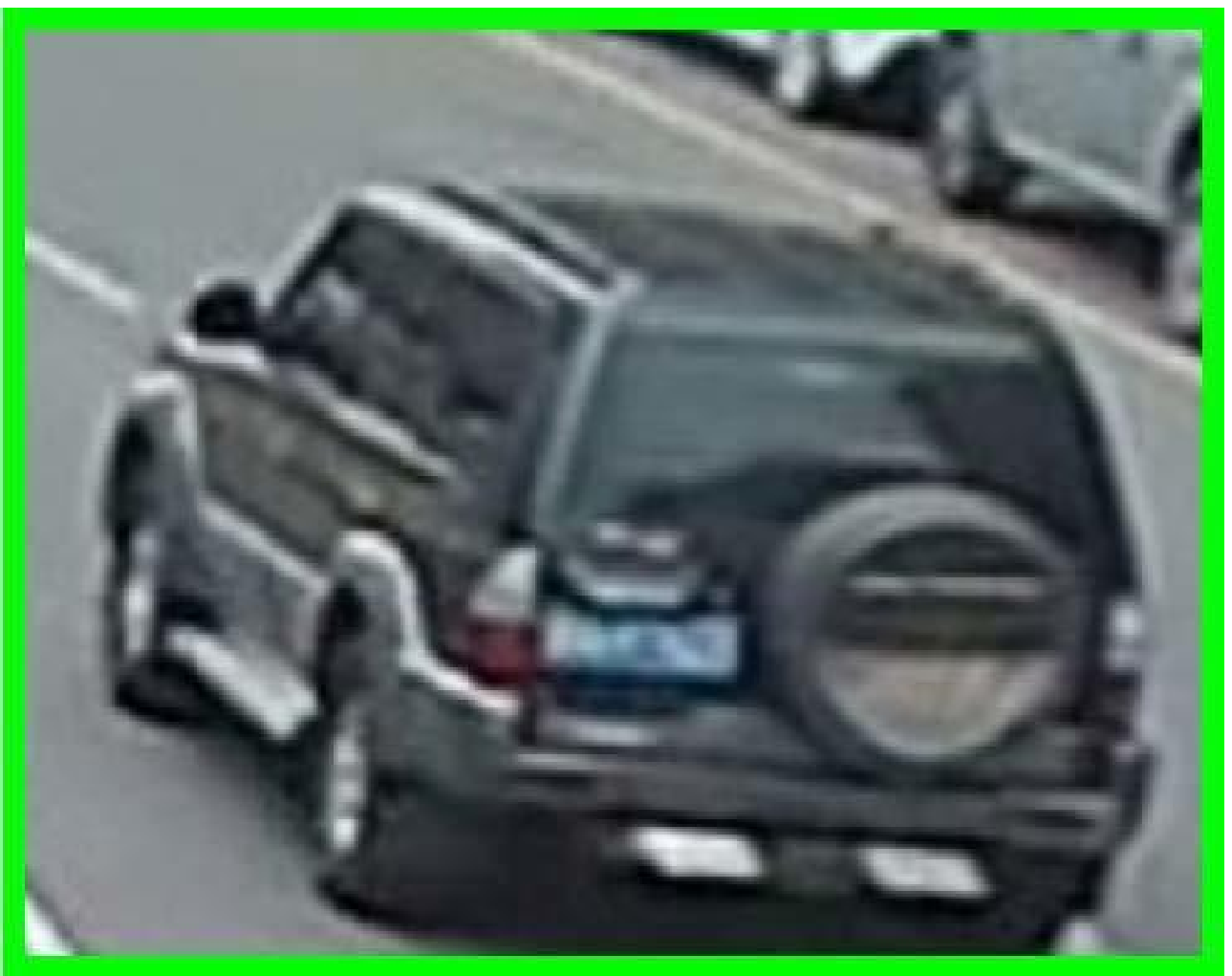}
                \caption{Rank $2$}
            \end{subfigure}
            \hspace{-0.2cm}
            &
            \begin{subfigure}[t]{0.08\textwidth}
                \includegraphics[width=\textwidth]{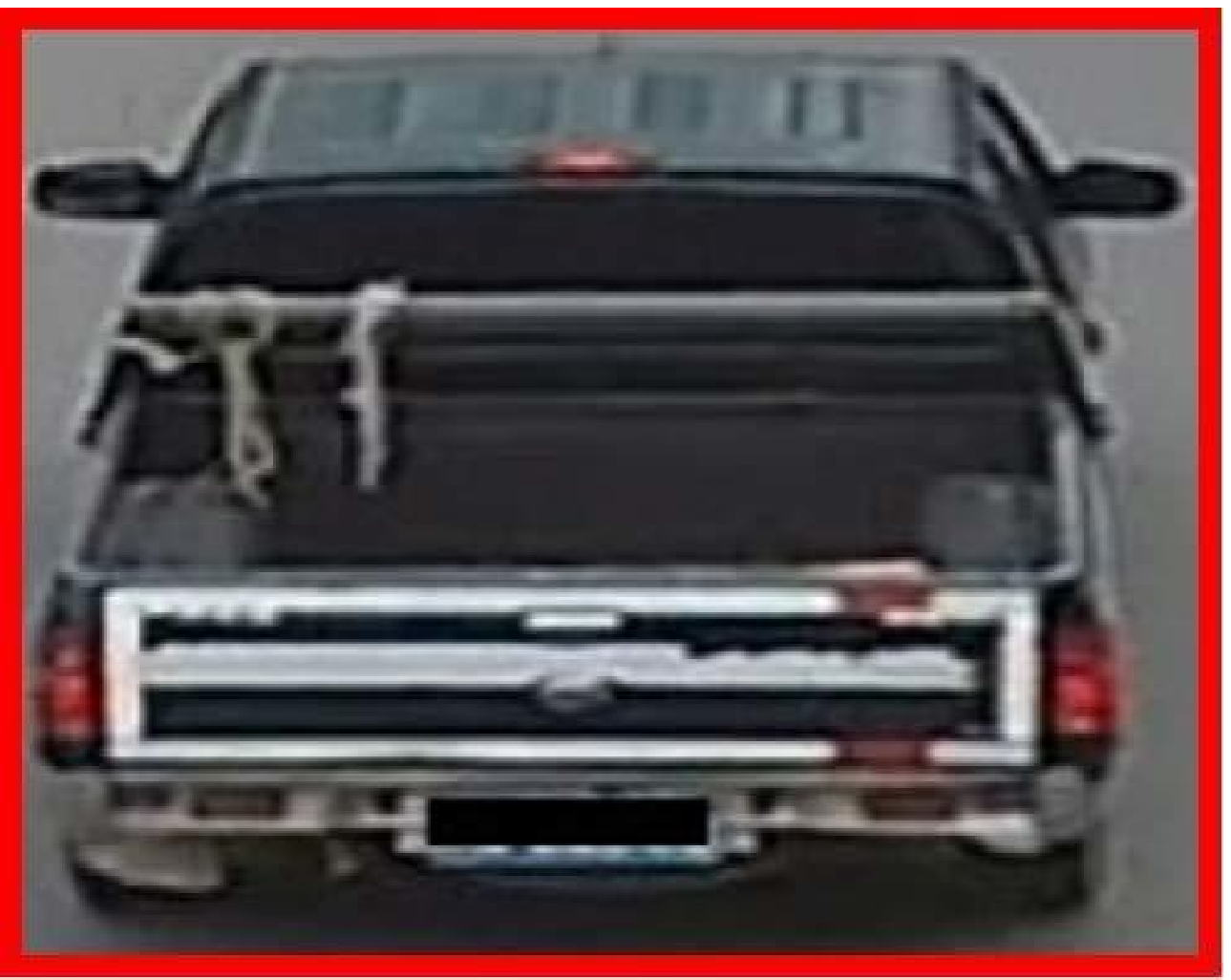}
                \caption{Rank $3$}
            \end{subfigure} 
            \\
            \begin{subfigure}[t]{0.08\textwidth}
                \includegraphics[width=\textwidth]{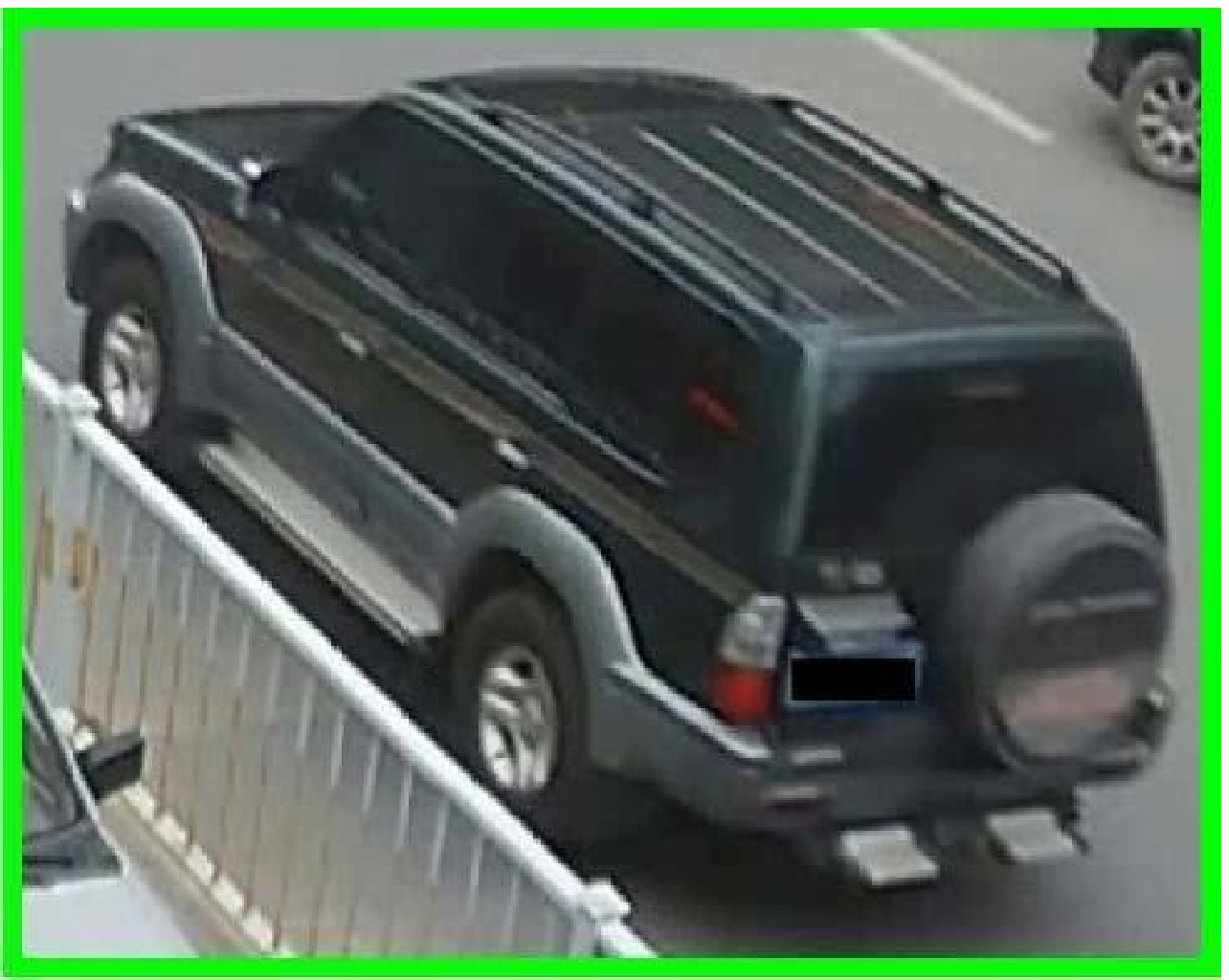}
                \caption{Rank $1$}
            \end{subfigure}
            \hspace{-0.2cm}
            &
            \begin{subfigure}[t]{0.08\textwidth}
                \includegraphics[width=\textwidth]{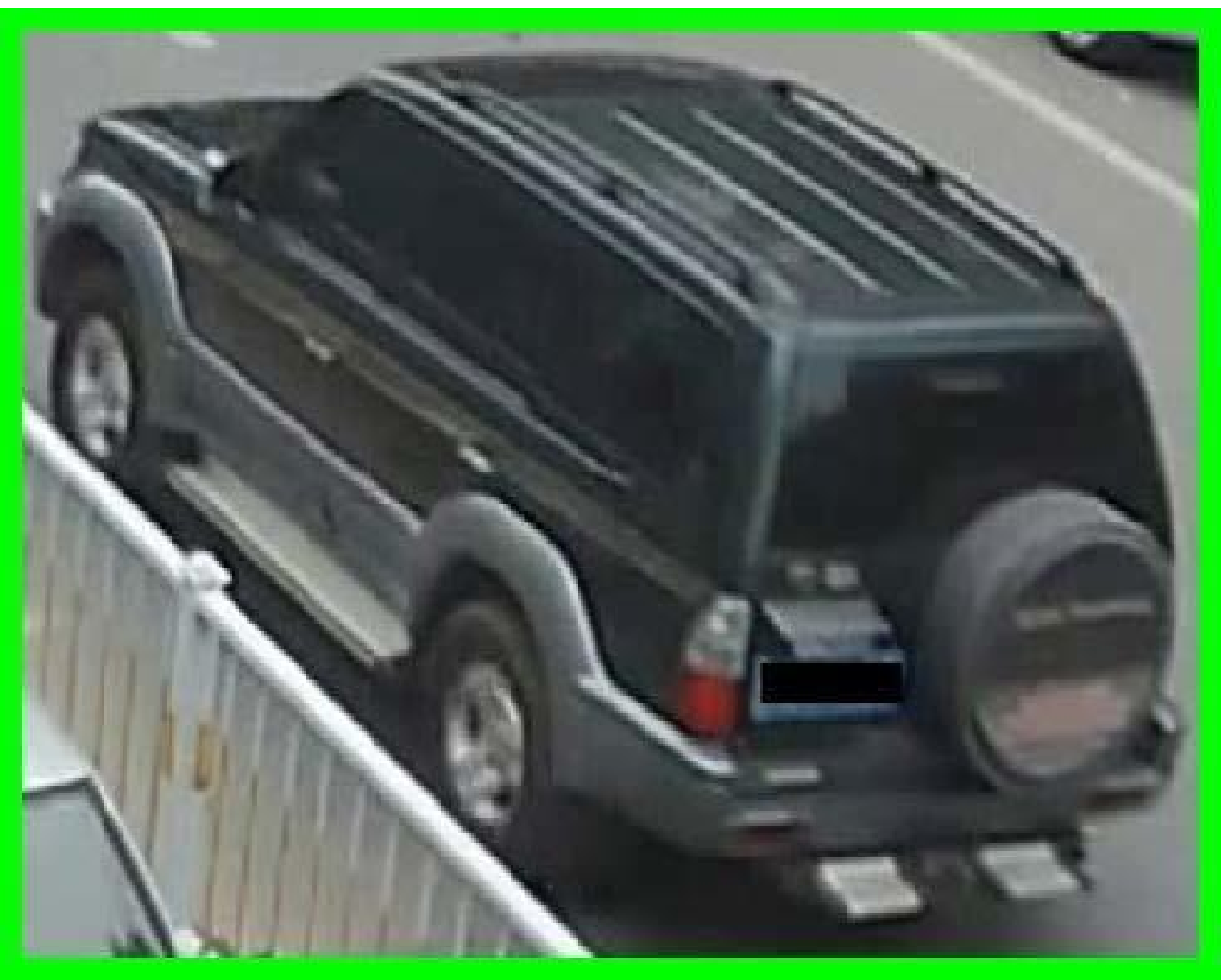}
                \caption{Rank $2$}
            \end{subfigure}
            \hspace{-0.2cm}
            &
            \begin{subfigure}[t]{0.08\textwidth}
                \includegraphics[width=\textwidth]{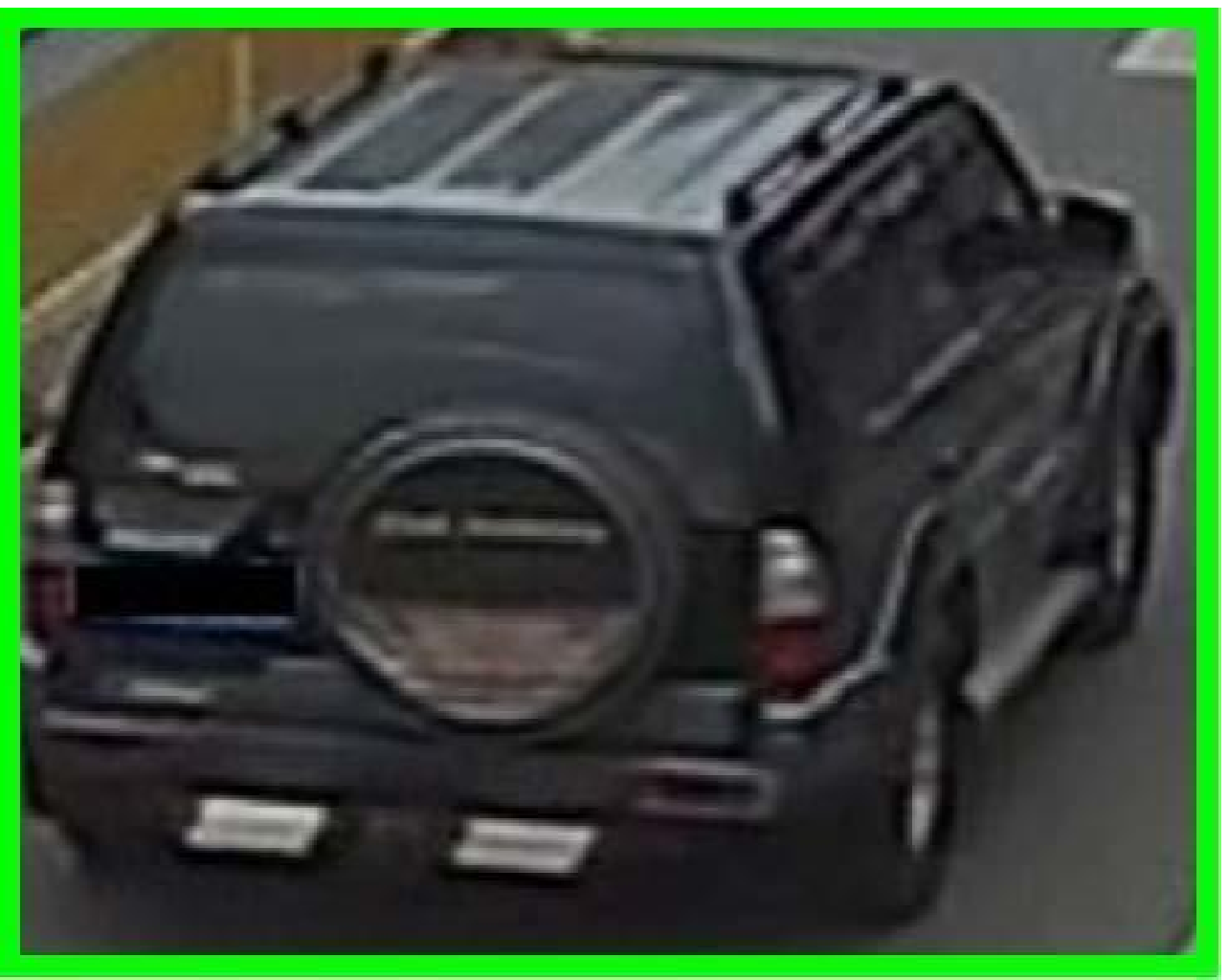}
                \caption{Rank $3$}
            \end{subfigure}
        \end{tabular}
    \end{tabular}
    \caption{Top three returned results of the baseline model (sub-figures b-d) versus the AAVER model (sub-figures e-g) on VeRi-776 dataset}
    \label{fig:baseline_VS_proposed_1}
\end{figure}

Figure \ref{fig:baseline_VS_proposed_1} plots the probe image and the top three returns of each baseline and the proposed model. It can be observed that AAVER significantly improves the performance over the baseline.

\subsection{Evaluation Results on VehicleID}

Images in this dataset have less variations in viewpoint, \emph{i.e.} mostly front and rear, compared to VeRi-776 dataset. For this dataset, the evaluation metrics are only CMC@1 and CMC@5 as there is only one true match in the gallery for each probe image. Table \ref{tab:baseline_proposed_comparison_Vehicle_ID} presents the re-identification results of baseline and the proposed models over test splits. As compared to baseline models, a significant increase in performance is observed when features from adaptive attention-based path are fused with global appearance features.

\begin{table}[!htb]
    \centering
    \caption{Performance comparison between baseline and proposed method on VehicleID dataset}
    \resizebox{\columnwidth}{!}{%
    \begin{tabular}{cc|c|c|c|c|}
    \cline{3-6}
    & & \multicolumn{2}{c|}{Baseline Model} & \multicolumn{2}{|c|}{AAVER Model}  \\
    \cline{2-6}
    & \multicolumn{1}{|c|}{Split} & ResNet-50 & ResNet-101 & \multicolumn{1}{|c|}{ResNet-50} & ResNet-101 \\
    
    \cline{2-6}
    \hline
    \multicolumn{1}{|c}{\multirow{3}{*}{CMC@1}} & \multicolumn{1}{|c||}{800} & 67.27 & 70.03 & \multicolumn{1}{|c|}{72.47} & \textbf{74.69} \\ \cline{2-6}
    \multicolumn{1}{ |c  }{}  &
    \multicolumn{1}{|c||}{1600}  & 62.03 & 65.26 & \multicolumn{1}{|c|}{66.85} & \textbf{68.62} \\ \cline{2-6}
    \multicolumn{1}{ |c  }{}  &
    \multicolumn{1}{|c||}{2400}  & 55.12 & 59.04 & \multicolumn{1}{|c|}{60.23} & \textbf{63.54} \\ \cline{1-6}
    \hline
    \hline
    \multicolumn{1}{|c}{\multirow{3}{*}{CMC@5}} & \multicolumn{1}{|c||}{800} & 89.05 & 89.81 & \multicolumn{1}{|c|}{93.22} & \textbf{93.82}  \\ \cline{2-6}
    \multicolumn{1}{ |c  }{}  &
    \multicolumn{1}{|c||}{1600}  & 84.31 & 84.96 & \multicolumn{1}{|c|}{89.39} & \textbf{89.95} \\ \cline{2-6}
    \multicolumn{1}{ |c  }{}  &
    \multicolumn{1}{|c||}{2400}  & 80.04 & 80.60 & \multicolumn{1}{|c|}{84.85} & \textbf{85.64} \\ \cline{1-6}
    \end{tabular}
    }
    \label{tab:baseline_proposed_comparison_Vehicle_ID}
\end{table}

\begin{figure}
    \centering
    \begin{tabular}[t]{cc}
        \begin{subfigure}{0.16\textwidth}
            \includegraphics[width=\textwidth]{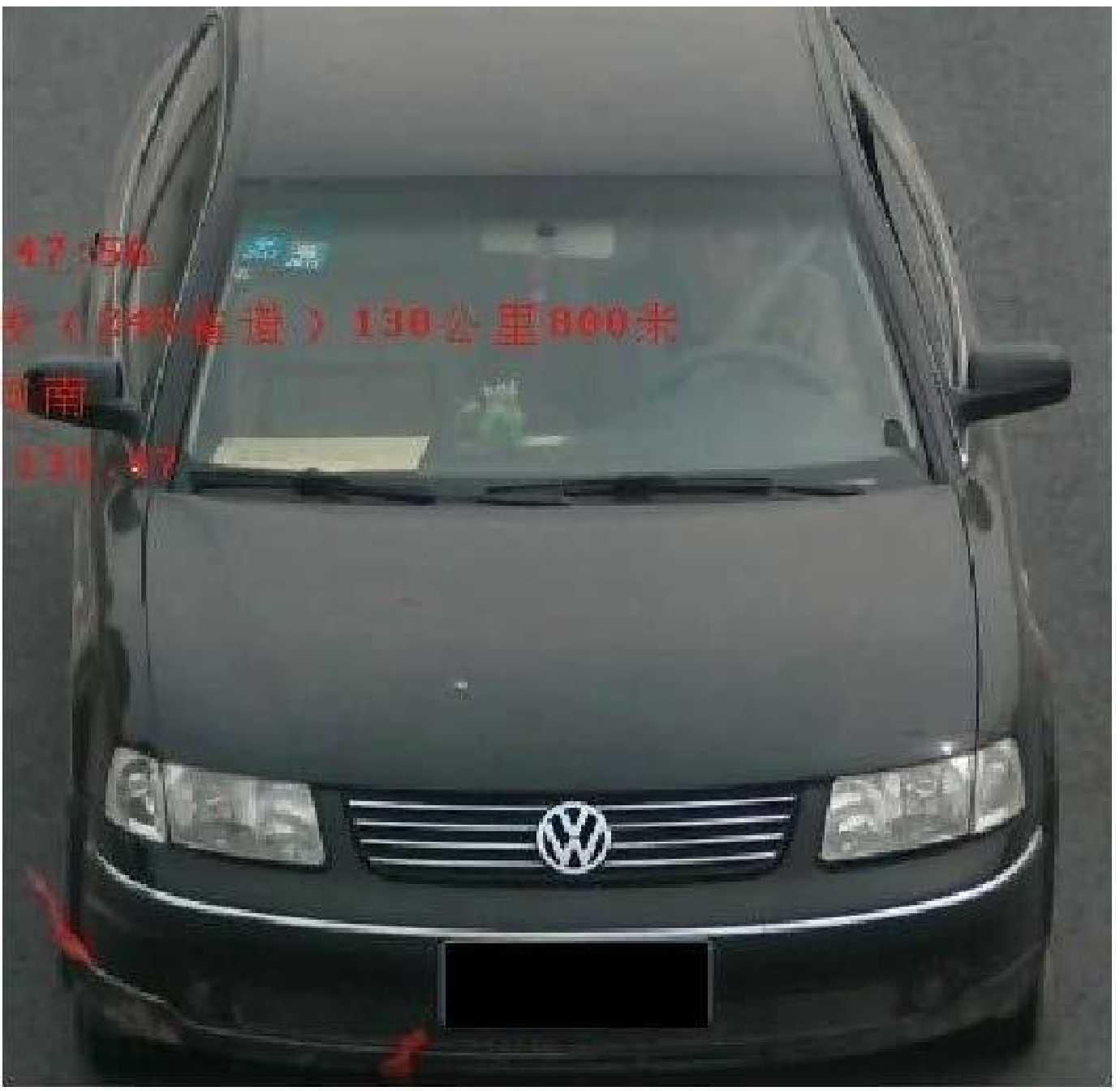}
            \caption{Probe Image}
        \end{subfigure}
        \hspace{-0.4cm}
        &  
        \begin{tabular}{ccc}
            \begin{subfigure}[t]{0.08\textwidth}
                \includegraphics[width=\textwidth]{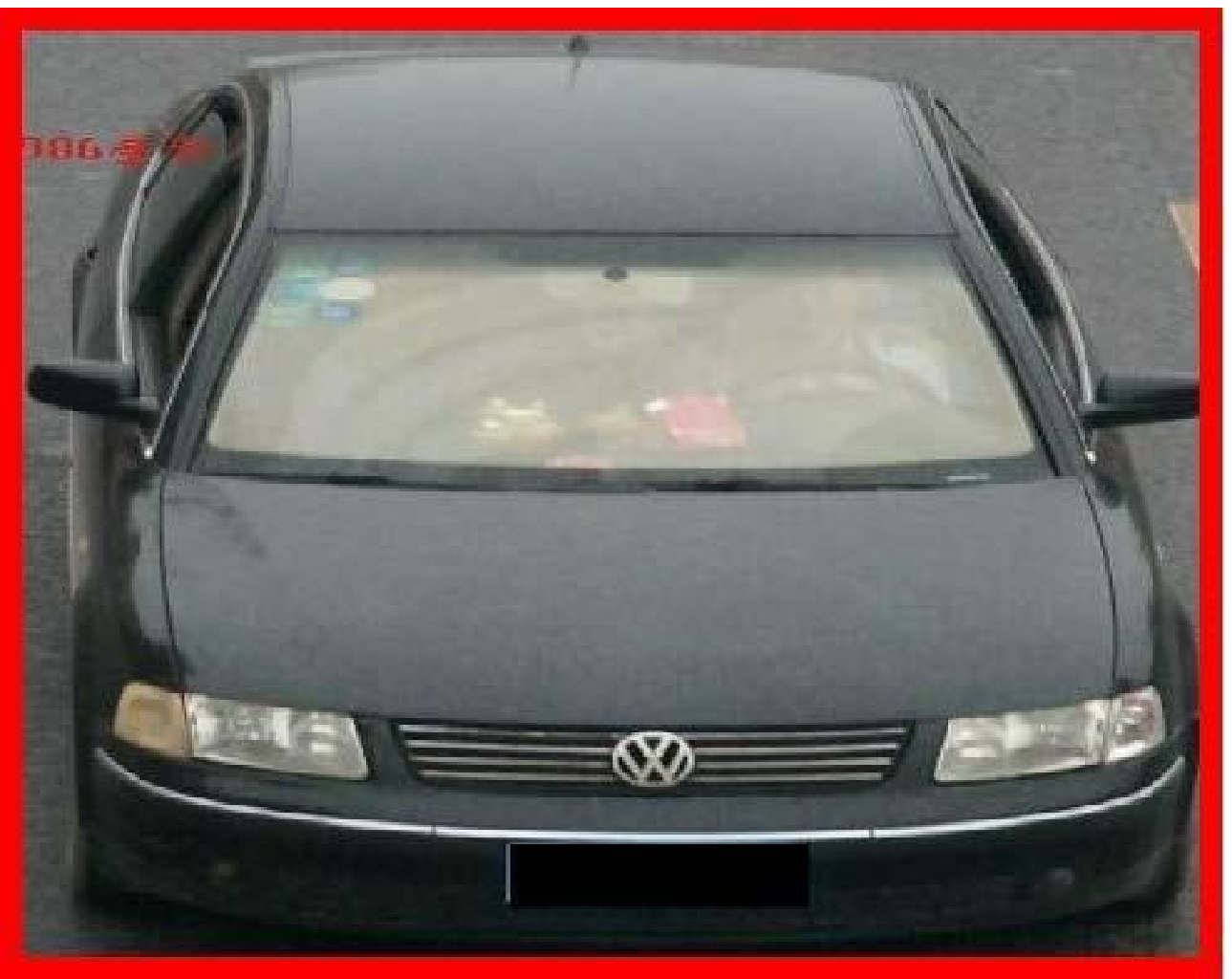}
                \caption{Rank $1$}
            \end{subfigure}
            \hspace{-0.1cm}
            &
            \begin{subfigure}[t]{0.08\textwidth}
                \includegraphics[width=\textwidth]{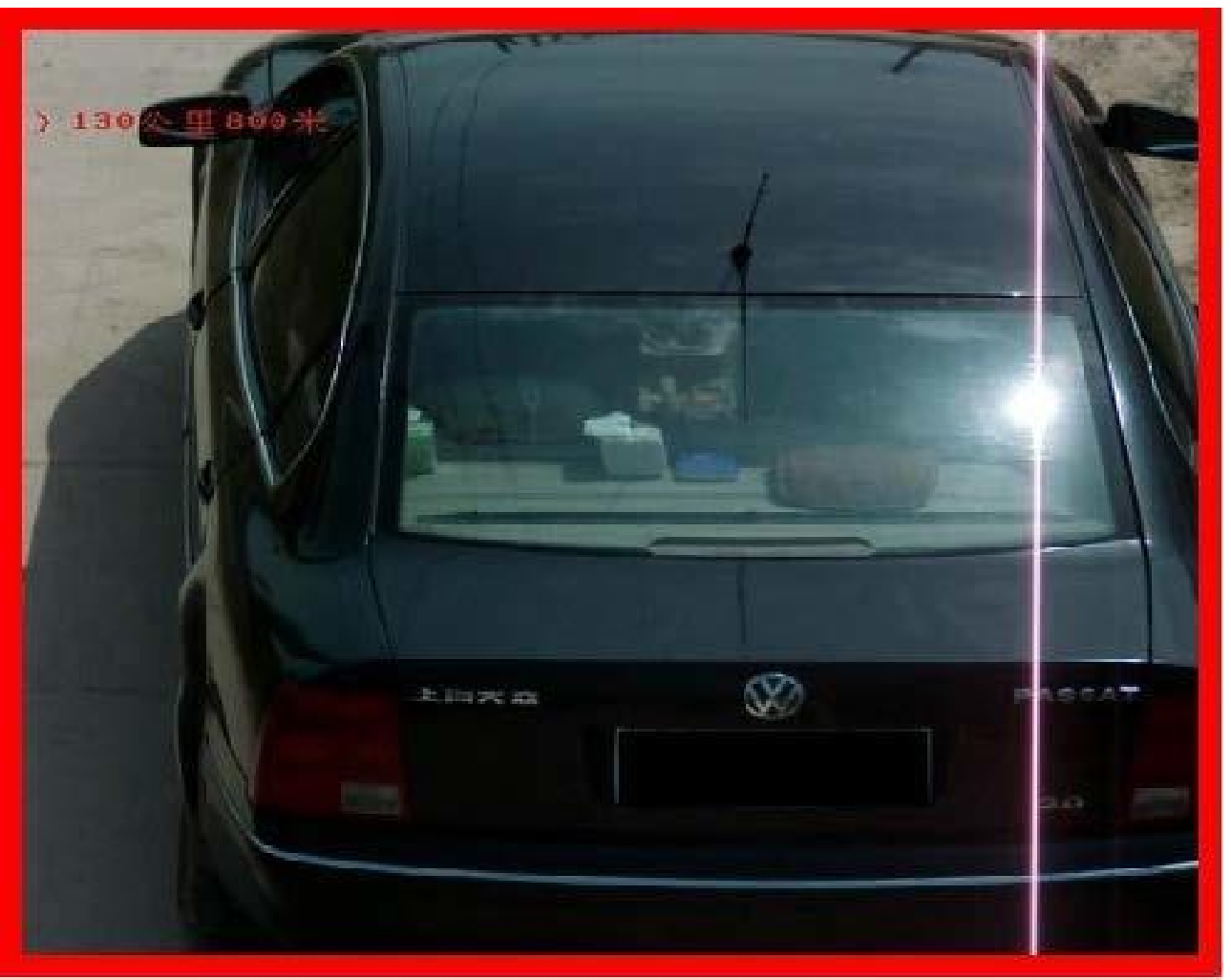}
                \caption{Rank $2$}
            \end{subfigure}
            \hspace{-0.1cm}
            &
            \begin{subfigure}[t]{0.08\textwidth}
                \includegraphics[width=\textwidth]{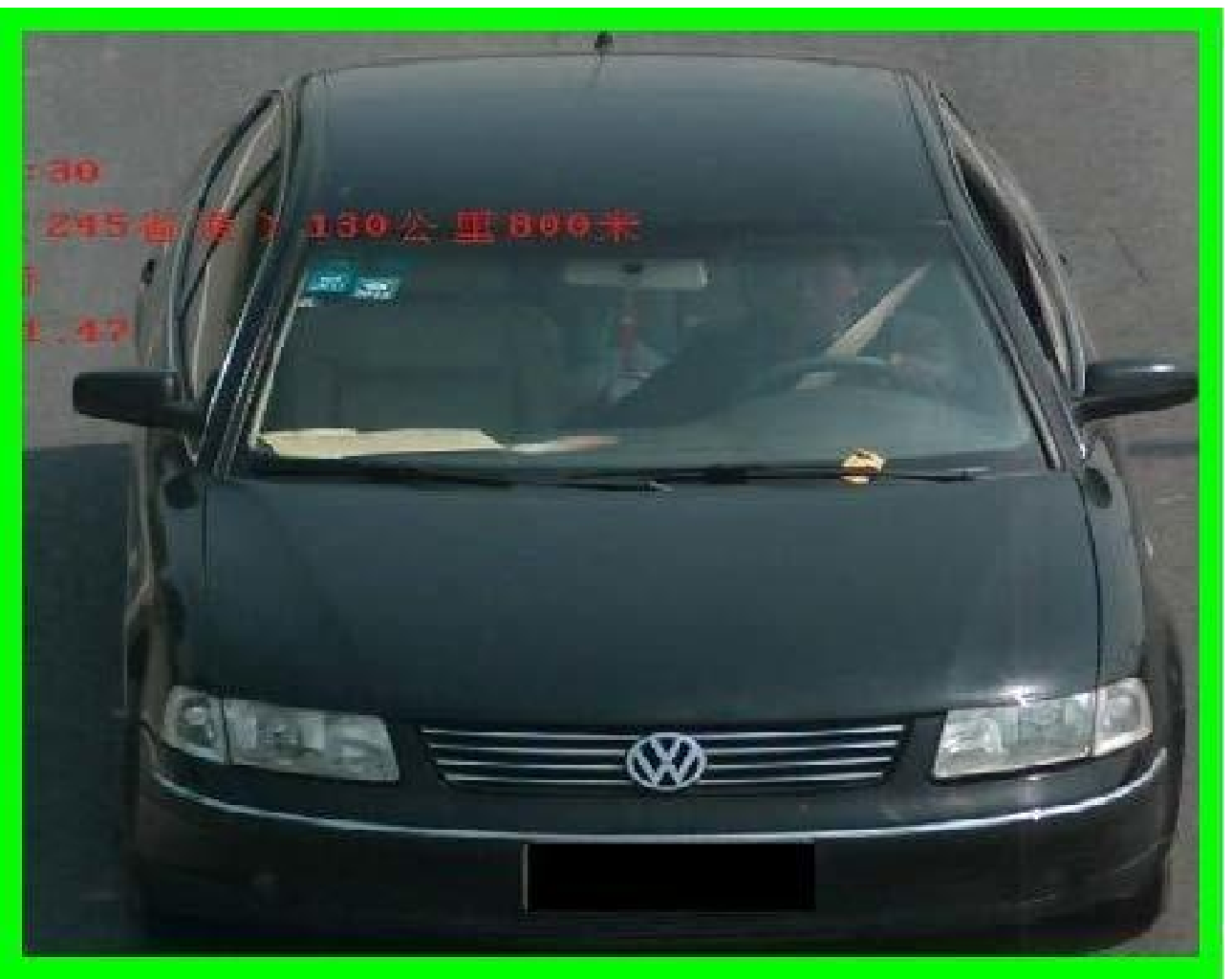}
                \caption{Rank $3$}
            \end{subfigure} 
            \\
            \begin{subfigure}[t]{0.08\textwidth}
                \includegraphics[width=\textwidth]{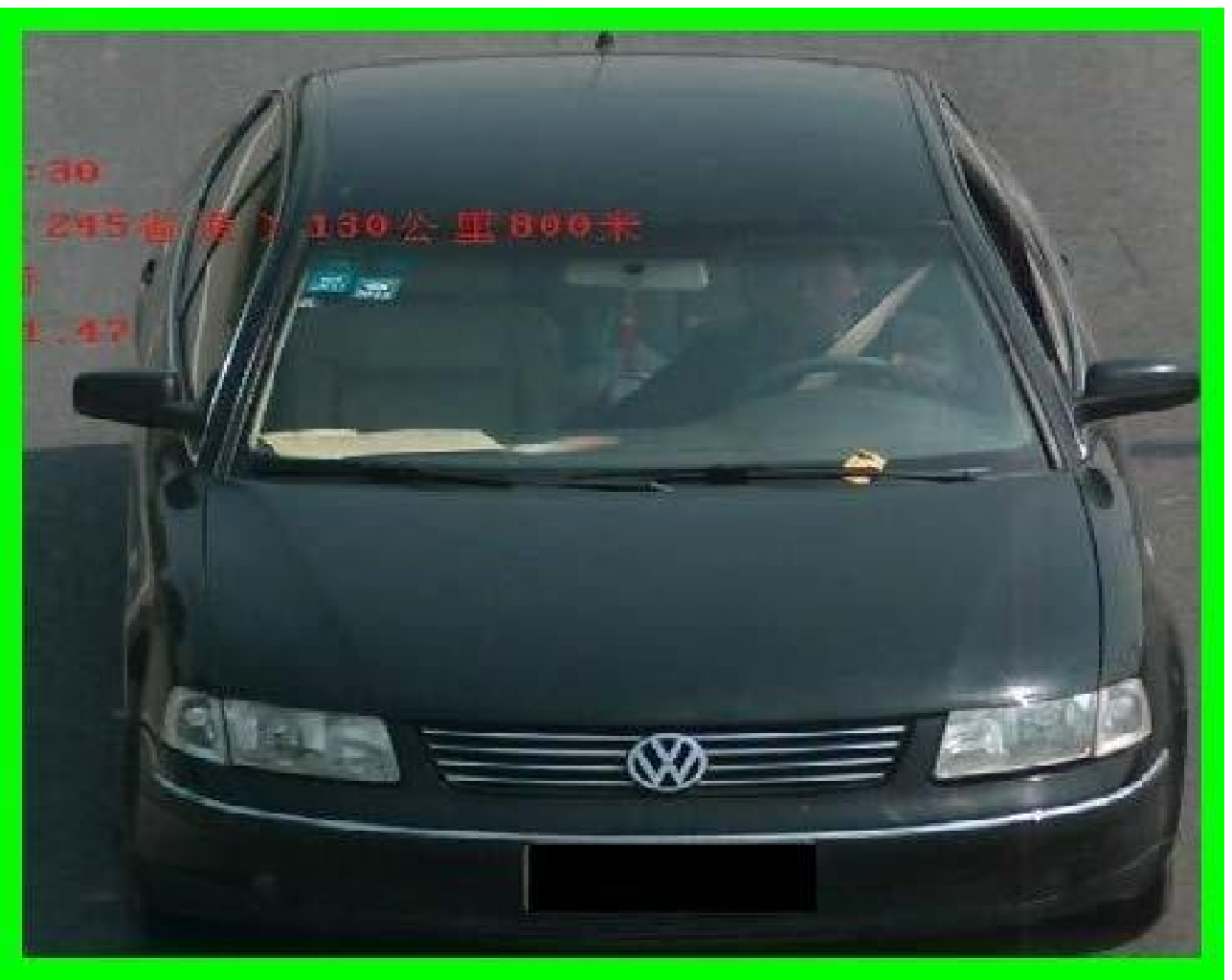}
                \caption{Rank$1$}
            \end{subfigure}
            \hspace{-0.1cm}
            &
            \begin{subfigure}[t]{0.08\textwidth}
                \includegraphics[width=\textwidth]{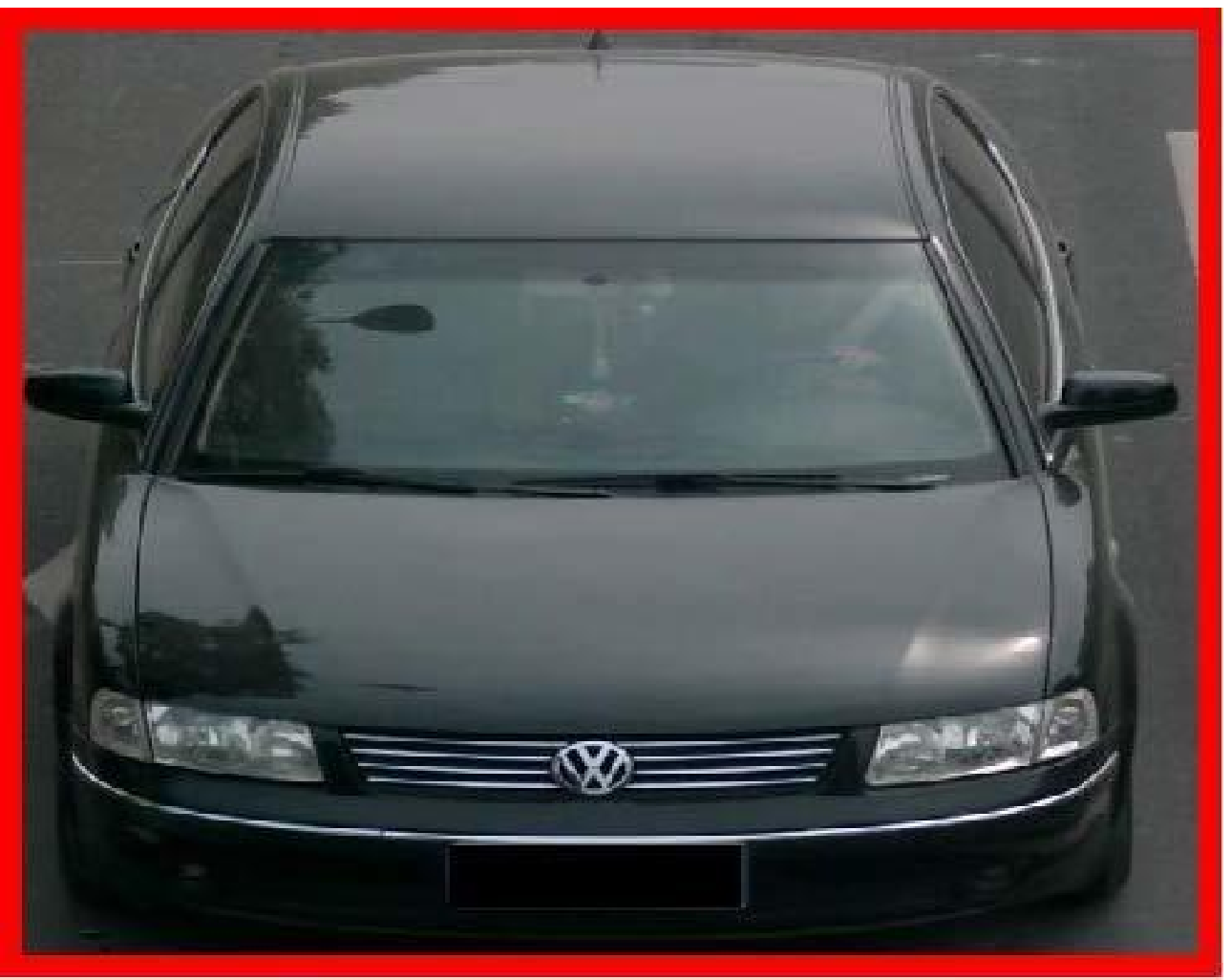}
                \caption{Rank $2$}
            \end{subfigure}
            \hspace{-0.1cm}
            &
            \begin{subfigure}[t]{0.08\textwidth}
                \includegraphics[width=\textwidth]{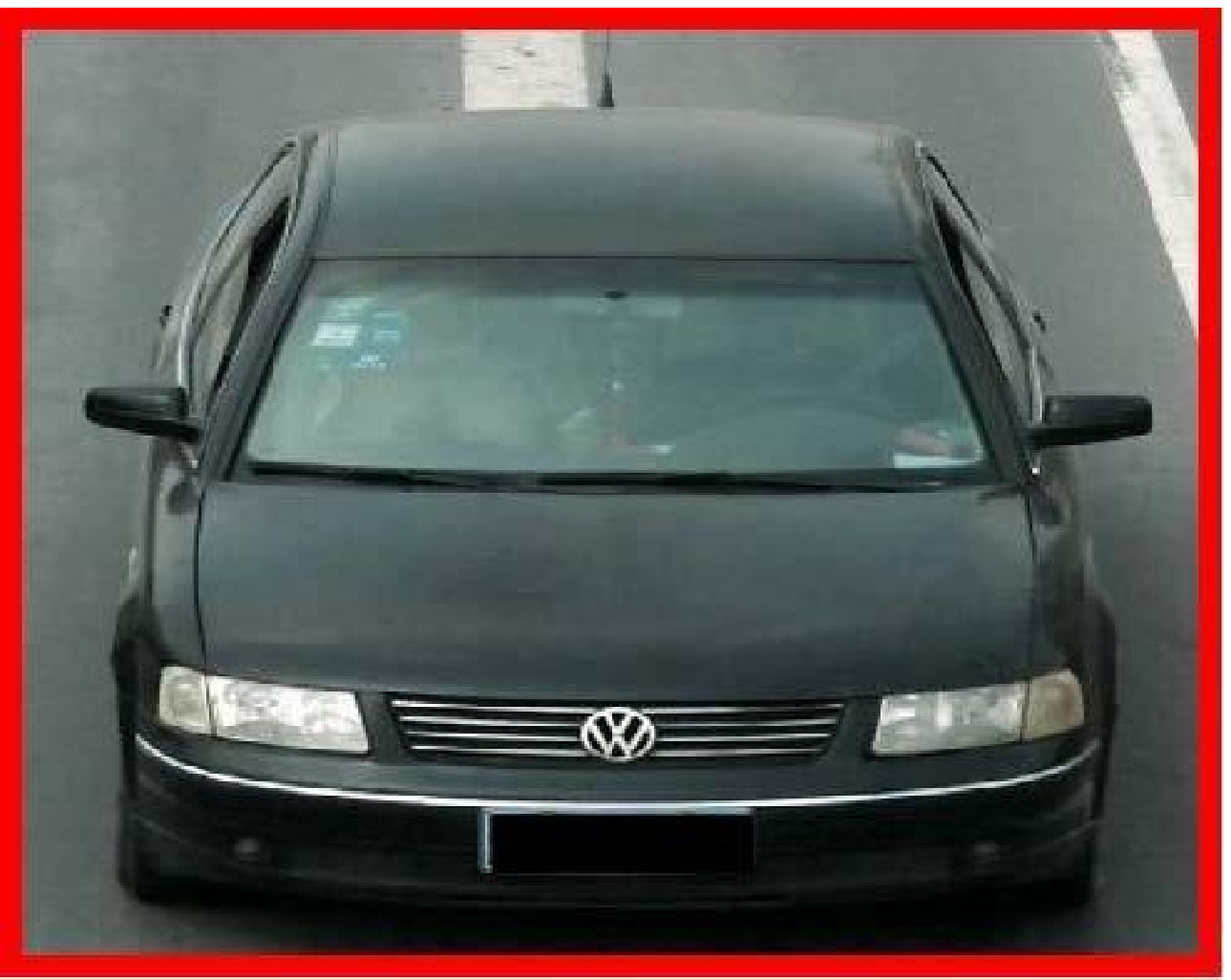}
                \caption{Rank $3$}
            \end{subfigure}
        \end{tabular}
    \end{tabular}
    \caption{Top three returned results of the baseline model (sub-figures b-d) versus the AAVER model (sub-figures e-g) on VehicleID dataset}
    \label{fig:baseline_VS_proposed_5}
\end{figure}

Figure \ref{fig:baseline_VS_proposed_5} shows an examples of a query from VehicleID dataset and the top three results returned by both global and adaptive attention model.

\subsection{Comparison with State of the Art Methods}

In this section, we compare AAVER model with ResNet-101 backbone against the recent state of the art methods. The results of this comparison are presented in Table \ref{tab:state_of_art_comp}.

\begin{table*}[t]
    \centering
    \caption{Comparison with recent methods and state of the arts}
    \label{tab:state_of_art_comp}
    \resizebox{2.1\columnwidth}{!}{%
    \begin{tabular}{c|c|c|cc|c|c|c|c|c|}
    \cline{2-10}
    & \multicolumn{9}{c|}{Dataset} \\
    \cline{2-10}
     & \multicolumn{3}{c}{\multirow{2}{*}{VeRi-776}} & \multicolumn{6}{||c|}{VehicleID} \\ \cline{5-10}
    & \multicolumn{3}{c}{} & \multicolumn{2}{||c|}{Test size = 800} & \multicolumn{2}{c|}{Test size = 1600} & \multicolumn{2}{c|}{Test size = 2400} \\
    \cline{1-10}
    \multicolumn{1}{|c|}{Method} & mAP & CMC@1 & CMC@5 & \multicolumn{1}{||c|}{CMC@1} & \multicolumn{1}{c|}{CMC@5} & CMC@1 & CMC@5 & CMC@1 & CMC@5 \\
    \hline
    \hline
    \multicolumn{1}{|c|}{SCPL \cite{shen2017learning}} & 58.27 & 83.49 & 90.04 & \multicolumn{1}{||c|}{-} & - & - & - & - & -  \\
    \hline
    \multicolumn{1}{|c|}{OIFE \cite{wang2017orientation}} & 48.00 & 65.9 & 87.7 & \multicolumn{1}{||c|}{-} & - & - & -  & - & - \\
    \hline
    \multicolumn{1}{|c|}{VAMI \cite{zhou2018viewpoint}} & 50.13 & 77.03 & 90.82 & \multicolumn{1}{||c|}{63.12} & 83.25 & 52.87 & 75.12 & 47.34 & 70.29 \\
    \hline
    \multicolumn{1}{|c|}{RAM \cite{liu2018ram}} & 61.5 & 88.6 & 94.0 & \multicolumn{1}{||c|}{\textbf{75.2}} & 91.5 & \textbf{72.3} & 87.0 & \textbf{67.7} & 84.5 \\
    \hline
    \hline
    \multicolumn{1}{|c|}{AAVER} & 61.18 &  88.97 & \textbf{94.70} & \multicolumn{1}{||c|}{74.69} & \textbf{93.82} & 68.62 & \textbf{89.95} & 63.54 & \textbf{85.64} \\
    \hline
    \multicolumn{1}{|c|}{AAVER + Re-ranking} & \textbf{66.35} &  \textbf{90.17} & 94.34 & \multicolumn{1}{||c|}{-} & - & - & - & - & - \\
    \hline
    \end{tabular}
    }
\end{table*}

 From Table \ref{tab:state_of_art_comp}, it can be observed that our proposed method is among the top performers of vehicle re-identification task and is the state of the art for most of the evaluation metrics on both VeRi-776 and VehicleID datasets. Note that in the absence of a deterministic test set for VehicleID dataset, one cannot provide the basis for a fair comparison. The reason lies in the fact that random gallery construction yields different evaluation results with relatively high variance even when averaged over ten repetitions. Finally, we have to emphasize on the necessity of using re-ranking as a post processing step whenever there are multiple instances of the probe image in the gallery. Here for the VeRi-776 dataset, re-ranking shows significant improvement and results in state of the art mAP and CMC@1 scores. Note that for VehicleID dataset re-ranking is not applicable as there is only one true match in the gallery for each probe image. 

\subsection{Ablation Studies}\label{subsec:ablation}

 We designed a set of experiments to study the impact of complementary information that the orientation conditioned branch provides. Note that in these experiments we only use the test split $800$ for the VehicleID dataset To this end, the following experiments have been conducted:

\begin{enumerate}
    \item In the first experiment we examined the depth of the layer in the global branch from which the global features are pooled and then fed to the orientation conditioned branch. To investigate this we tried pooling features after Res2, Res3 and Res4 blocks of spatial size of $56\times56$, $28\times28$ and $14\times14$. Table \ref{tab:which_layer_to_pool} demonstrates the results of this experiment. It can be observed that as we go from shallow to deeper layers,
    the features become more abstract and focusing on parts of deep feature maps do not help in providing a robust representation of vehicles with minute differences.

    \begin{table}[h]
    \centering
    \caption{Experiment 1: Depth of pooled global features}
    \resizebox{\columnwidth}{!}{%
    \begin{tabular}{|c|c|c|c|c|}
    \cline{1-5}
    Dataset & features size & mAP & CMC@1 & CMC@5  \\
    \hline
    \multirow{3}{*}{VeRi-776} & $56\times56$ & \textbf{0.612} & \textbf{88.97} & \textbf{94.70}   \\
    \cline{2-5}
     & $28\times28$ & 0.608 & 88.50 & 94.58 \\
    \cline{2-5}
     & $14\times14$ & 0.597 & 85.88 & 93.03 \\
    \hline
    \hline
    \multirow{3}{*}{VehicleID} & $56\times56$ & - & \textbf{74.69} & \textbf{93.82}  \\
    \cline{2-5}
     & $28\times28$ & - & 72.60 & 93.24 \\
    \cline{2-5}
     & $14\times14$ & - & 71.09 & 92.13 \\
    \hline
    
    \end{tabular}
    }
    \label{tab:which_layer_to_pool}
    \end{table}
    
    \item In our method, we use two streams for extracting global and local features from vehicle images, so we were keen to see whether a single branch can extract discriminative features that encompass global as well as local differences. To test this hypothesis, instead of pooling features from the global branch we fused the selected heatmaps into the global branch by concatenation and used the output as the representation for a vehicle image.
    Table \ref{tab:single_vs_two_stream} depicts the result of this experiment, for both VeRi and VehicleID datasets. we can infer that the re-identification performance drops
    significantly by relying on a single branch.
    
    \begin{table}[h]
    \centering
    \caption{Experiment 2: Single versus Dual-branch feature extraction}
    \begin{tabular}{|c|c|c|c|c|}
    \cline{1-5}
    Dataset & Type & mAP & CMC@1 & CMC@5  \\
    \hline
    \multirow{2}{*}{VeRi-776} & Single & 0.528 & 80.93 & 90.52  \\
    \cline{2-5}
     & Dual & \textbf{0.612} & \textbf{88.97} & \textbf{94.70} \\
    \hline
    \hline
    \multirow{2}{*}{VehicleID} & Single & - & 69.61 & 91.45  \\
    \cline{2-5}
     & Dual & - & \textbf{74.69} & \textbf{93.82} \\
    \hline
    \end{tabular}
    \label{tab:single_vs_two_stream}
    \end{table}
    \vspace{-0.2cm}
    \item In the final set of experiments we scrutinize the way in which the information from vehicle key-point heatmaps are incorporated in the proposed model. Our work is in some aspects similar to \cite{wang2017orientation} which groups a fixed set of key-points and combines all the corresponding heatmaps into one map by adding them together. Therefore, we conduct this experiment under the same settings as of \cite{wang2017orientation}. Table \ref{tab:keypoint_heatmap_incorporation}
    shows the results of these experiments. The type "Combined" in Table \ref{tab:keypoint_heatmap_incorporation} refers to the method in \cite{wang2017orientation}. We can conclude that using all heatmaps combined into one group does not result in competitive results as the adaptive selection of heatmaps. 
    This validates the hypothesis that not all the key-points contribute to a discriminative representation of the vehicle.
    
    \begin{table}[h]
    \centering
    \caption{Experiment 3: Key-points heatmaps utilization}
    \resizebox{\columnwidth}{!}{%
    \begin{tabular}{|c|c|c|c|c|}
    \cline{1-5}
    Dataset & Type & mAP & CMC@1 & CMC@5  \\
    \hline
    \hline
    \multirow{2}{*}{VeRi-776} & Combined\cite{wang2017orientation} & 0.606 & 87.66 & 94.17\\
    \cline{2-5}
    & AAVER & \textbf{0.612} & \textbf{88.97} & \textbf{94.70} \\
    \hline
    \hline
    \multirow{2}{*}{VehicleID} & Combined\cite{wang2017orientation} & - & 71.79 & 92.10\\
    \cline{2-5}
     & AAVER & - & \textbf{74.69} & \textbf{93.82} \\
    \hline
    \end{tabular}
    }
    \label{tab:keypoint_heatmap_incorporation}
    \end{table}
\end{enumerate}

%% file: conclusion.tex
\section{Conclusions and Future Work}\label{sec:conclusion}
In this paper, we present a robust end-to-end framework for state of the art vehicle re-identification. We present a dual path model AAVER which combines macroscopic global features with localized discriminative features to efficiently identify a probe image in a gallery of varying sizes. In addition, we establish benchmarks for key-point detection and orientation prediction on VeRi-776 dataset. Lastly, we advocate for the adoption of re-ranking when considering the performance of future vehicle re-identification methods. Adaptive key-point selection conditioned on vehicle orientation is vital for discriminating between vehicles of the same make, model and color. Evaluating on both VeRi-776 and VehicleID shows the strength of our proposed method. Lastly, we conduct an ablation study to understand the influence of the adaptive key-point selection step.

In the future, we plan to extend our key-point module to align vehicle images to a canonical coordinates system before comparing a given pair of images. Similarly, we can learn a 3D representation of vehicles to be used in other tasks such as vehicular speed estimation.

%% file: acknowledgement.tex
\section{Acknowledgement}
\small{This research is supported in part by the Northrop Grumman Mission Systems Research in Applications for Learning Machines (REALM) initiative, It is also supported in part by the Office of the Director of National Intelligence (ODNI), Intelligence Advanced Research Projects Activity (IARPA), via IARPA R\&D Contract No. D17PC00345. The views and conclusions contained herein are those of the authors and should not be interpreted as necessarily representing the official policies or endorsements, either expressed or implied, of ODNI, IARPA, or the U.S. Government. The U.S. Government is authorized to reproduce and distribute reprints for Governmental purposes notwithstanding any copyright annotation thereon.}

%% file: supp.tex
\section*{Supplementary Material}
\noindent
In this section we present the performance of our method on the newly released VeRi-Wild dataset\cite{lou2019large}. This dataset is collected in a network of 174 surveillance cameras covering a large urban area and captures unconstrained scenarios. It is composed of 416,314 images (277,797/138,517 for train/test sets) of 40,671 different vehicle identities. The test set, similar to VehicleID dataset is divided into three Small (41,861), Medium (69,389) and Large (138,517) images splits. The evaluation for this dataset follows the same protocol as VeRi-776. Table \ref{tab:baseline_proposed_comparison_VeRi-Wild} summarizes the result of our baseline and proposed AAVER model. 

\begin{table}[hbt!]
    \centering
    \caption{Performance comparison between baseline and proposed method on VeRi-Wild dataset}
    \resizebox{1.05\columnwidth}{!}{%
    \begin{tabular}{cc|c|c|c|c|}
    \cline{3-6}
    & & \multicolumn{2}{c|}{Baseline Model} & \multicolumn{2}{|c|}{AAVER Model}  \\
    \cline{2-6}
    & \multicolumn{1}{|c|}{Split} & ResNet-50 & ResNet-101 & \multicolumn{1}{|c|}{ResNet-50} & ResNet-101 \\
    
    \cline{2-6}
    \hline
    \multicolumn{1}{|c}{\multirow{3}{*}{mAP}} & \multicolumn{1}{|c||}{Small} & 60.22 & 60.99 & \multicolumn{1}{|c|}{60.62} & \textbf{62.23} \\ \cline{2-6}
    \multicolumn{1}{ |c  }{}  &
    \multicolumn{1}{|c||}{Medium}  & 51.21 & 52.49 & \multicolumn{1}{|c|}{51.77} & \textbf{53.66} \\ \cline{2-6}
    \multicolumn{1}{ |c  }{}  &
    \multicolumn{1}{|c||}{Large}  & 38.89 & 38.99 & \multicolumn{1}{|c|}{40.42} & \textbf{41.68} \\ \cline{1-6}
    \hline
    \hline
    \multicolumn{1}{|c}{\multirow{3}{*}{CMC@1}} & \multicolumn{1}{|c||}{Small} & 71.37 & 72.97 & \multicolumn{1}{|c|}{74.60} & \textbf{75.80} \\ \cline{2-6}
    \multicolumn{1}{ |c  }{}  &
    \multicolumn{1}{|c||}{Medium}  & 62.84 & 65.02 & \multicolumn{1}{|c|}{65.76} & \textbf{68.24} \\ \cline{2-6}
    \multicolumn{1}{ |c  }{}  &
    \multicolumn{1}{|c||}{Large}  & 52.00 & 54.65 & \multicolumn{1}{|c|}{56.03} & \textbf{58.69} \\ \cline{1-6}
    \hline
    \hline
    \multicolumn{1}{|c}{\multirow{3}{*}{CMC@5}} & \multicolumn{1}{|c||}{Small} & 90.10 & 91.57 & \multicolumn{1}{|c|}{91.60} & \textbf{92.70}  \\ \cline{2-6}
    \multicolumn{1}{ |c  }{}  &
    \multicolumn{1}{|c||}{Medium}  & 86.58 & 87.30 & \multicolumn{1}{|c|}{87.16} & \textbf{88.88} \\ \cline{2-6}
    \multicolumn{1}{ |c  }{}  &
    \multicolumn{1}{|c||}{Large}  & 77.48 & 79.52 & \multicolumn{1}{|c|}{79.67} & \textbf{81.59} \\ \cline{1-6}
    \end{tabular}
    }
    \label{tab:baseline_proposed_comparison_VeRi-Wild}
\end{table}

From Table \ref{tab:baseline_proposed_comparison_VeRi-Wild} it can be seen that for all splits of VeRi-Wild dataset like VeRi-776 and VehicleID datasets, a significant boost is obtained by conditioning the features on the vehicle's orientation and corresponding key-points. 

\begin{figure}[H]
    \centering
    \begin{tabular}[t]{cc}
        \begin{subfigure}{0.16\textwidth}
            \includegraphics[width=\textwidth]{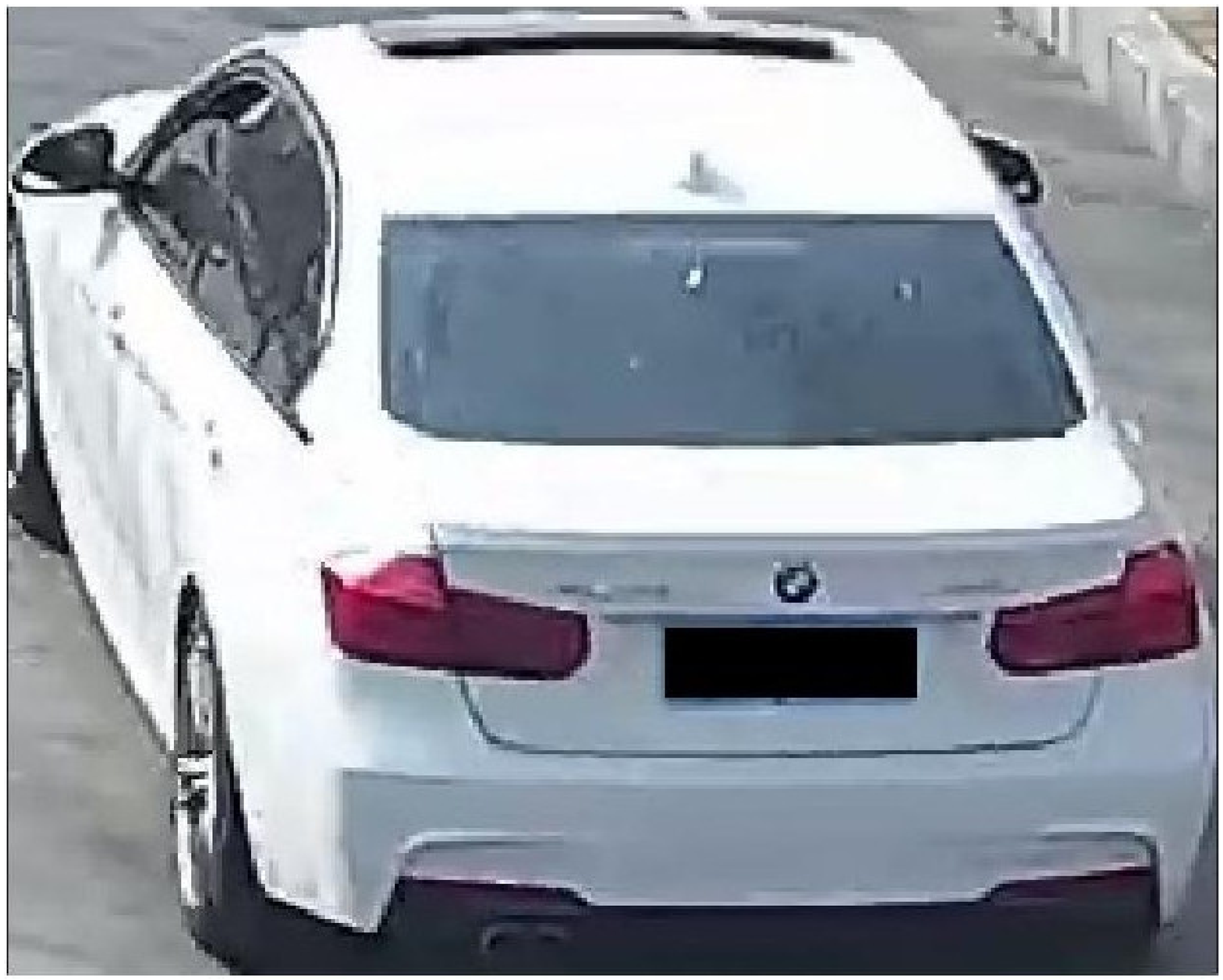}
            \caption{Probe Image}
        \end{subfigure}
        \hspace{-0.4cm}
        &  
        \begin{tabular}{ccc}
            \begin{subfigure}[t]{0.08\textwidth}
                \includegraphics[width=\textwidth]{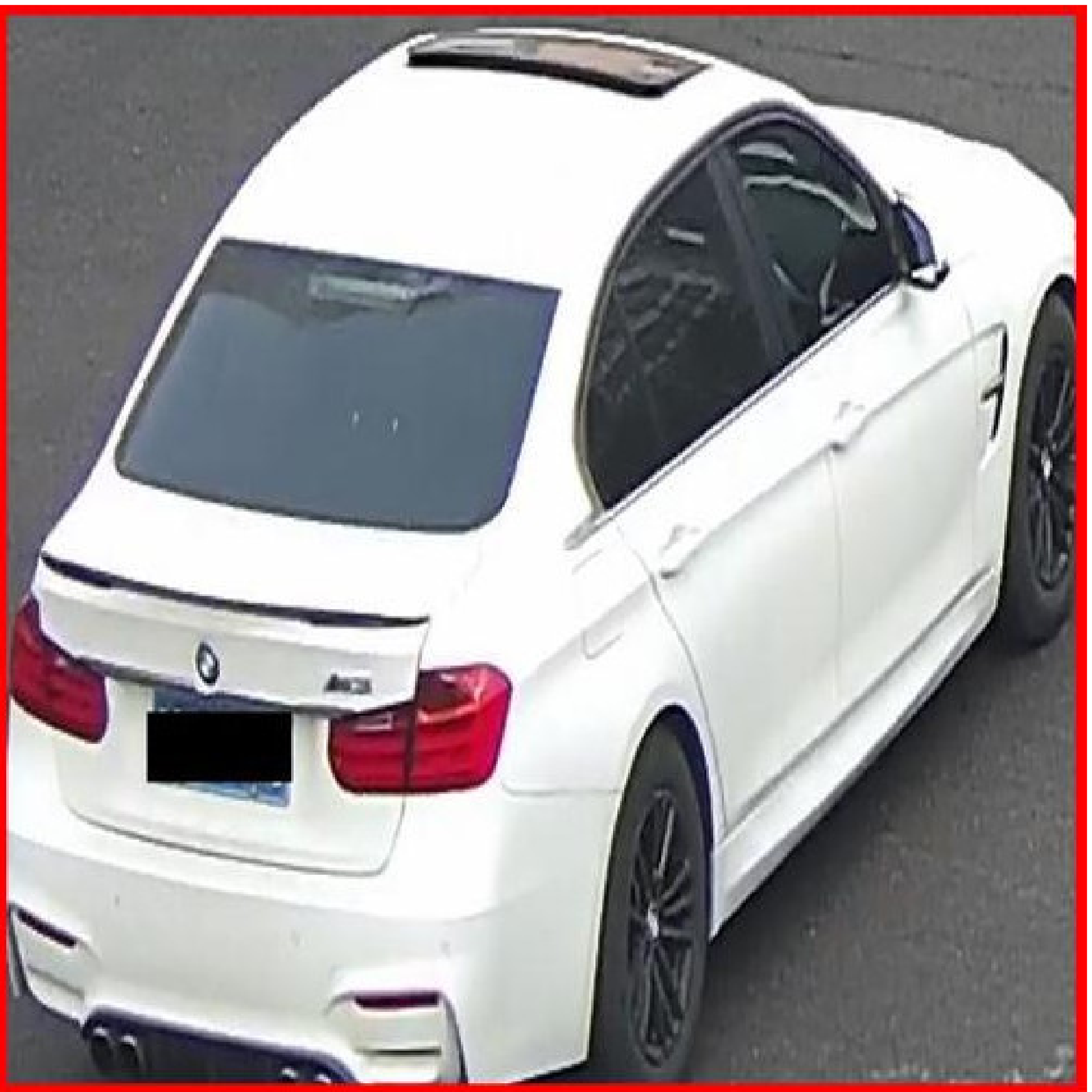}
                \caption{Rank $1$}
            \end{subfigure}
            \hspace{-0.1cm}
            &
            \begin{subfigure}[t]{0.08\textwidth}
                \includegraphics[width=\textwidth]{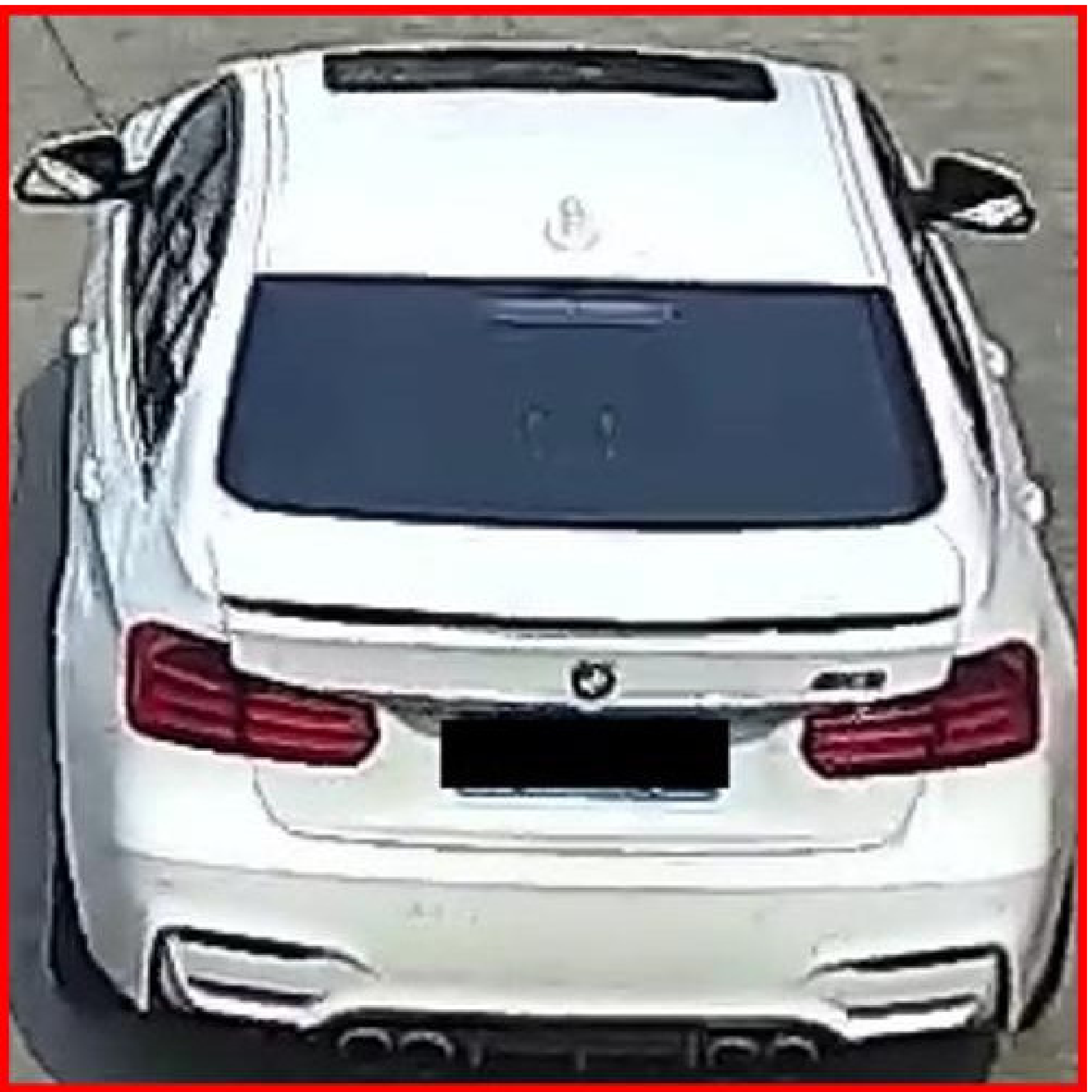}
                \caption{Rank $2$}
            \end{subfigure}
            \hspace{-0.1cm}
            &
            \begin{subfigure}[t]{0.08\textwidth}
                \includegraphics[width=\textwidth]{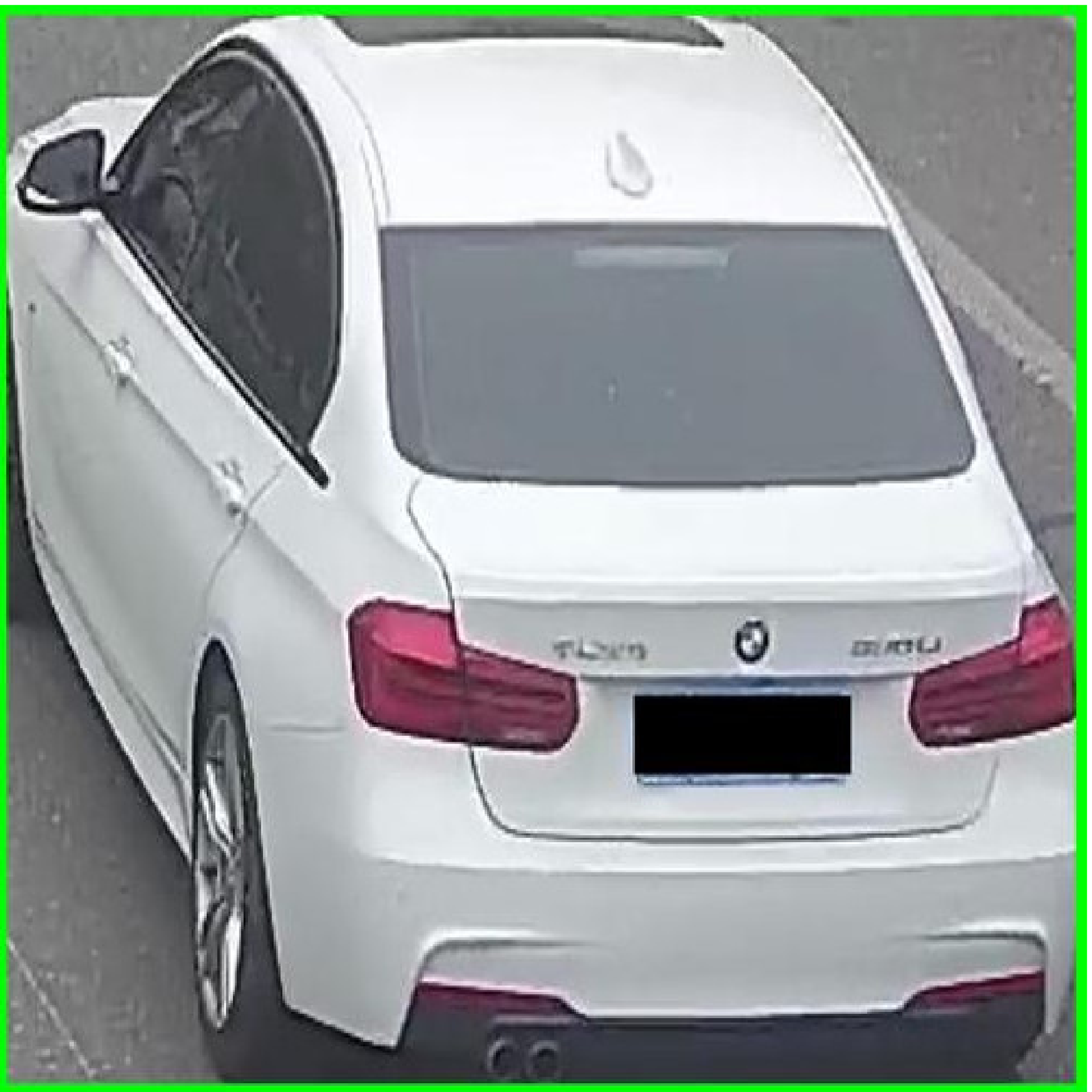}
                \caption{Rank $3$}
            \end{subfigure} 
            \\
            \begin{subfigure}[t]{0.08\textwidth}
                \includegraphics[width=\textwidth]{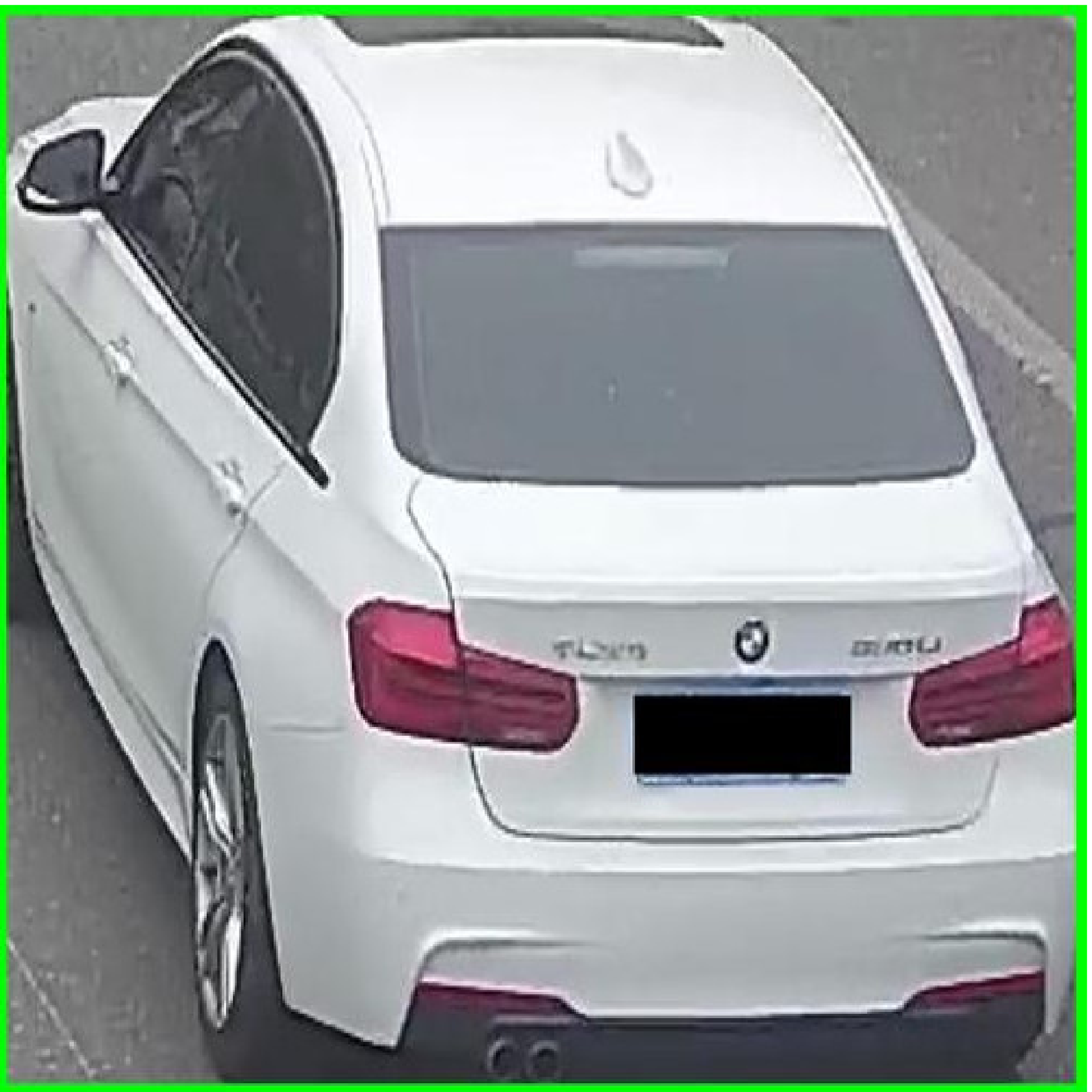}
                \caption{Rank$1$}
            \end{subfigure}
            \hspace{-0.1cm}
            &
            \begin{subfigure}[t]{0.08\textwidth}
                \includegraphics[width=\textwidth]{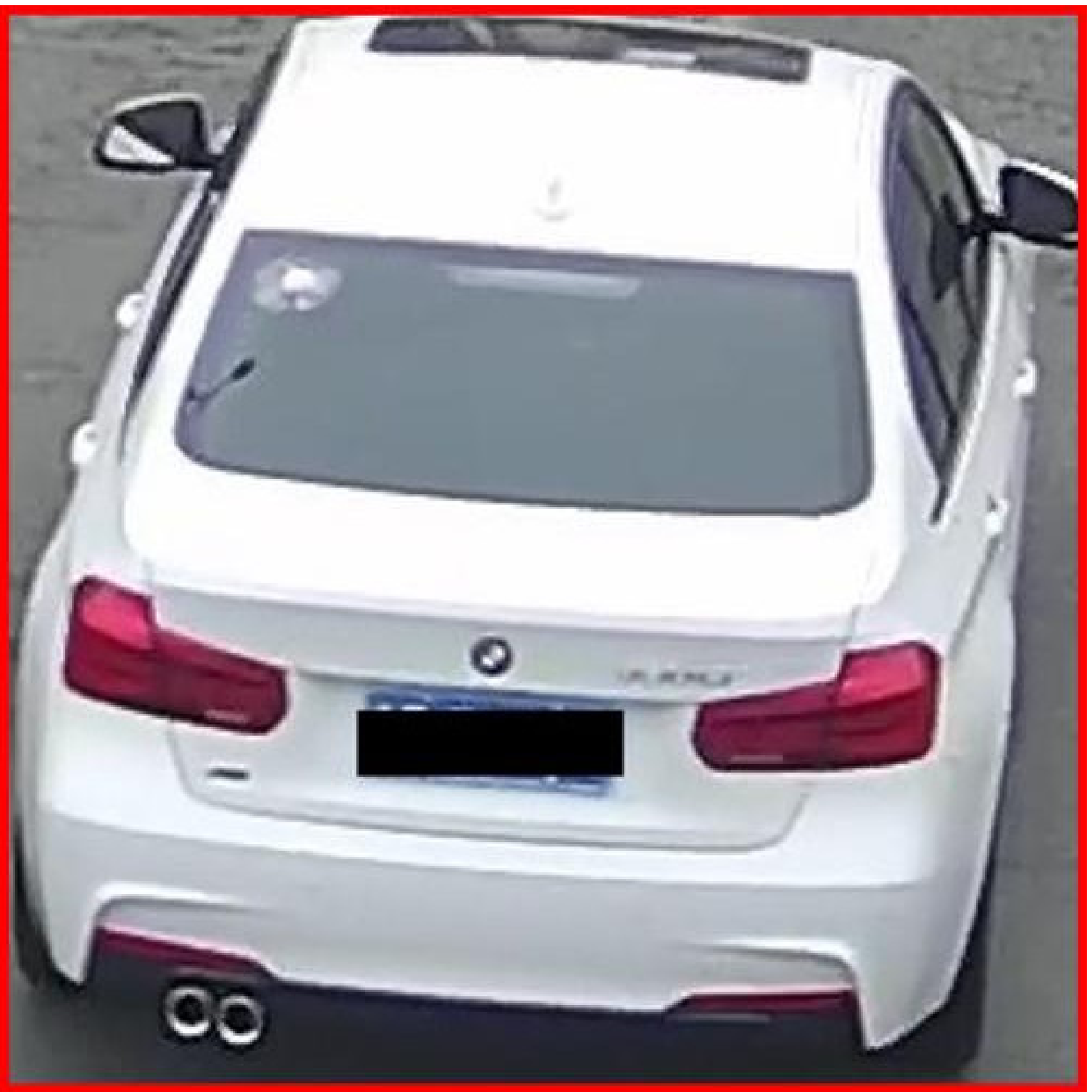}
                \caption{Rank $2$}
            \end{subfigure}
            \hspace{-0.1cm}
            &
            \begin{subfigure}[t]{0.08\textwidth}
                \includegraphics[width=\textwidth]{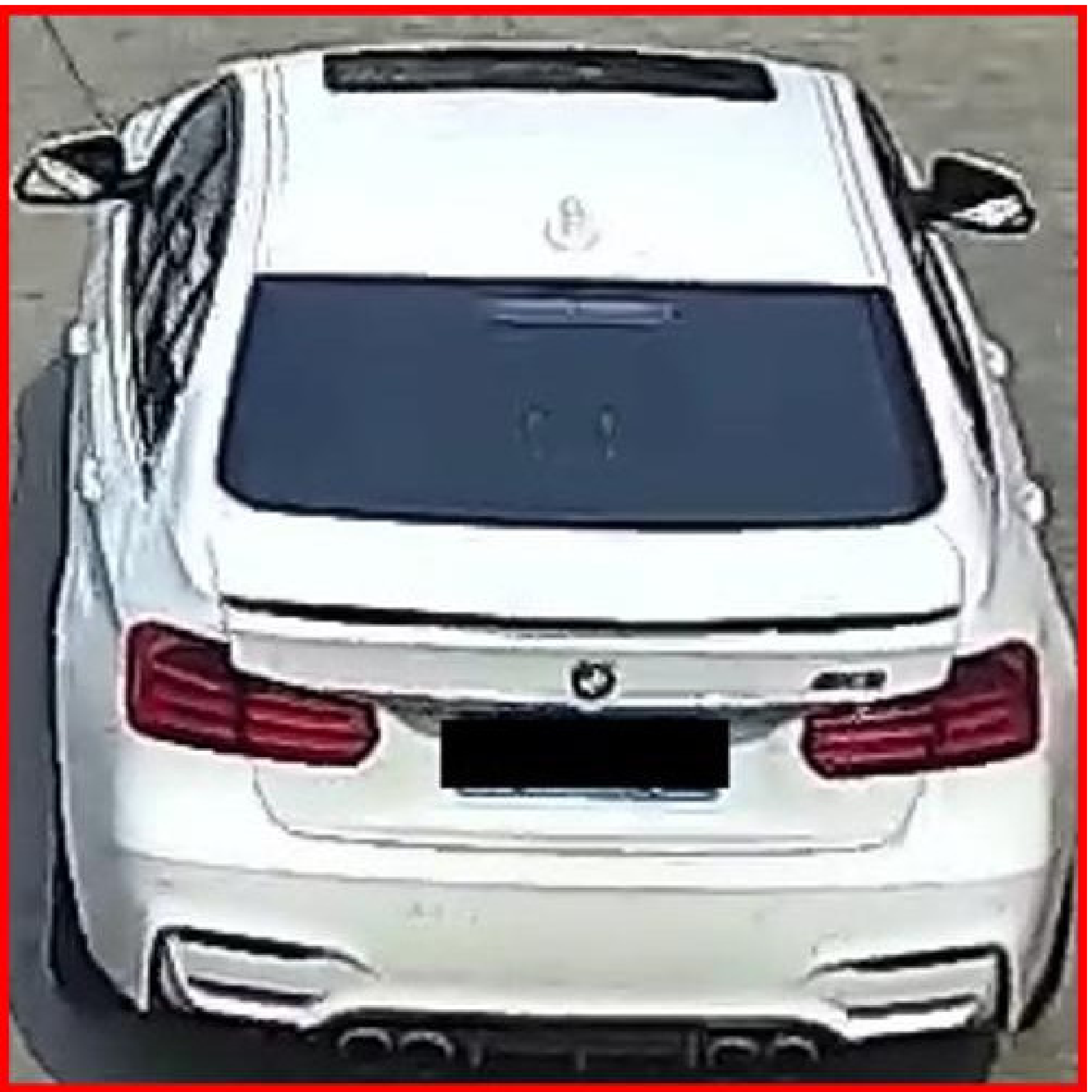}
                \caption{Rank $3$}
            \end{subfigure}
        \end{tabular}
    \end{tabular}
    \caption{Top three returned results of the baseline model (sub-figures b-d) versus the AAVER model (sub-figures e-g) on VeRi-Wild dataset}
    \label{fig:baseline_VS_proposed_6}
\end{figure}

Figure \ref{fig:baseline_VS_proposed_6} shows an examples of a query from VeRi-Wild dataset and the top three results returned by both global and adaptive attention models.
